\documentclass{article}

\PassOptionsToPackage{numbers, compress}{natbib}

     \usepackage[preprint]{neurips_2021}

\usepackage[utf8]{inputenc} 
\usepackage[T1]{fontenc}    
\usepackage{hyperref}       
\usepackage{url}            
\usepackage{array,booktabs}       
\usepackage{amsfonts}       
\usepackage{nicefrac}       
\usepackage{microtype}      
\usepackage{xcolor}         
\usepackage{verbatim}

\usepackage{colortbl}
\usepackage{graphicx}
\usepackage{subfigure}
\usepackage{amsthm}
\usepackage{amsmath}
\usepackage{amssymb}
\usepackage{blkarray}
\usepackage{mathtools}
\usepackage{multirow}
\usepackage{bm}
\usepackage{enumitem}
\usepackage{comment}
\usepackage{listings}
\usepackage{algorithm}
\usepackage{cleveref}
\usepackage{picinpar}
\usepackage{wrapfig}
\usepackage{xcolor}
\usepackage[figuresright]{rotating}
\usepackage{lipsum}
\usepackage{algpseudocode}
\usepackage{algorithmicx}

\renewcommand{\epsilon}{\varepsilon}
\renewcommand{\tilde}{\widetilde}

\def\:#1{\protect \ifmmode {\mathbf{#1}} \else {\textbf{#1}} \fi}

\renewcommand{\epsilon}{\varepsilon}

\newcolumntype{M}[1]{>{\centering\arraybackslash}m{#1}}

\DeclarePairedDelimiterX{\inp}[2]{\langle}{\rangle}{#1, #2}

\newcommand\crule[3][black]{\textcolor{#1}{\rule{#2}{#3}}}

\providecommand*\algorithmautorefname{Algorithm}

\title{Self-Supervised Graph Learning with Proximity-based Views and Channel Contrast}

\author{%
  Wei Zhuo \\
  Sun Yat-sen University\\
  \texttt{zhuow5@mail2.sysu.edu.cn} \\
   \And
   Guang Tan \\
   Sun Yat-sen University \\
   \texttt{tanguang@mail.sysu.edu.cn} \\
}

\begin{document}

\maketitle

\begin{abstract}

We consider graph representation learning in a self-supervised manner. Graph neural networks (GNNs) use neighborhood aggregation as a core component that results in feature smoothing among nodes in proximity. While successful in various prediction tasks, such a paradigm falls short of capturing nodes' similarities over a long distance, which proves to be important for high-quality learning. To tackle this problem, we strengthen the graph with two additional graph views, in which nodes are directly linked to those with the most similar features or local structures. Not restricted by connectivity in the original graph, the generated views allow the model to enhance its expressive power with new and complementary perspectives from which to look at the relationship between nodes. Following a contrastive learning approach, we propose a method that aims to maximize the agreement between representations across generated views and the original graph. We also propose a channel-level contrast approach that greatly reduces computation cost, compared to the commonly used node level contrast, which requires computation cost quadratic in the number of nodes. Extensive experiments on seven assortative graphs and four disassortative graphs demonstrate the effectiveness of our approach.  

\end{abstract}

\section{Introduction}
Graph neural networks (GNNs) have emerged as a powerful tool for many graph analytic problems such as node classification~\cite{kipf2017semi, velickovic2018graph, hamilton2017inductive}, graph classification~\cite{ying2018hierarchical}, link prediction~\cite{zhang2018link} and community detection~\cite{bruna2017community}. Most of these tasks are semi-supervised, requiring a certain number of labels to guide the learning process. In reality, the labeling information is sometimes difficult to obtain, or may not conform to the overall distribution of the data, resulting in inaccurate models. This has motivated the self-supervised graph learning approaches that perform well without relying on label information, and thus have gained increasing interest in the community.


Most GNNs~\cite{kipf2017semi,hamilton2017inductive, velickovic2018graph} depend on the assumption that similar nodes are likely to be connected and belong to the same class. Through local feature smoothing by neighborhood aggregation, these models tend to generate node embeddings that preserve the proximity of nodes from the original graph. Such an approach, however, faces challenges in some real-world networks, where neighborhoods do not necessarily mean similarity~\cite{zhu2020beyond}, and sometimes truly similar nodes are far apart. For example, in social networks, the unique attributes of celebrities would be diluted if aggregated with their followers. To enhance their unique attributes, the nodes need to aggregate messages from those with similar attributes, regardless of distance. This requires the GNN model to preserve the proximity of nodes in the {\em feature space} instead of in the original graph. In still other cases, it is found that structural similarity plays a particularly important role for graph learning. For example, for catalysts in the protein-protein interaction (PPI) network of a cell~\cite{ribeiro2017struc2vec}, nodes with similar structural features are defined to be similar, establishing a notion of proximity in the {\em topology space}. With supervision signals or prior knowledge, most GNNs use only specific types of proximity information. For example, AM-GCN~\cite{wang2020gcn} and SimP-GCN~\cite{jin2021node} try to leverage proximity in the feature space and the original graph; Struc2vec~\cite{ribeiro2017struc2vec} and RolX~\cite{henderson2012rolx} preserve proximity in topology space, while ignoring proximity information in other forms. 



In this paper, we argue that the three types of proximity, namely proximity in the original graph, in the feature space, and in the topology space, provide complementary views of the graph, and thus combining all of them in a proper way can significantly improve the robustness and adaptability of GNNs. By exploiting the inherent consistency between different graph views, we are able to learn useful embeddings without label information. Our method follows the self-supervised contrastive learning (CL) approach~\cite{wu2018unsupervised,he2020momentum,chen2020simple}. By contrasting positive pairs against negative pairs in different views of the graph, CL has demonstrated clear advantages on representation learning~\cite{velickovic2018deep, hassani2020contrastive,zhu2020deep} and pre-training~\cite{you2020graph,qiu2020gcc} tasks on graphs. 

\paragraph{Our Contributions} We present the \textsc{\textbf{F}eature \& \textbf{T}opology proximity preserving \textbf{G}raph \textbf{CL}} (FT-GCL), a CL based unsupervised graph representation learning (GRL) framework. First, we reconstruct the graph in the feature space and topology space to generate two graph views, in which similar nodes in terms of features or local topology are linked. In the contrastive stage, our model alternately maximizes an agreement between the representations of a generated view and the original graph, with positive and negative pairs at the channel level instead of the node level. Compared with the state-of-the-art method, GRACE~\cite{zhu2020deep}, which requires computation complexity $O(N^2)$, where $N$ is the number of nodes, FT-GCL reduces the number of contrastive pairs to $O(d^2)$, where $d$ is the output dimension. 
Our theoretical analysis from the perspective of maximizing mutual information confirms its rationality. We conduct experiments node classification task on a total of 11 real-world benchmark datasets including 7 assortative graphs and 4 disassortative graphs, and demonstrate that FT-GCL and its variants outperform representative unsupervised graph learning methods and sometimes defeat semi-supervised methods.

\section{Related Work}
{\bf Self-supervised learning on graphs} aims to construct supervision signals from the graph itself without external labels. 
Earlier methods based on shallow neural networks construct supervision signals from graph structures to enforce the representations of nodes in the same local context to be similar, where local context can be random walk sequences~\cite{perozzi2014deepwalk,grover2016node2vec}, specific order neighbors~\cite{tang2015line} or community members~\cite{wang2017community}. With the success of graph neural networks, some methods use multilayer graph autoencoder to learn to reconstruct certain parts of the graph, where the parts of the graph can be the adjacency matrix~\cite{2016variational,pan2018adversarially}, node features~\cite{park2019symmetric} or both nodes and edges~\cite{hu2020gpt}. Recently, Contrastive learning (CL) has been successfully applied in graph representation learning. The key idea is to contrast views by maximizing agreement between positive pair and disagreement between negative pairs, where contrastive pairs can be subgraph pairs~\cite{qiu2020gcc}, subgraph-graph pairs~\cite{you2020graph}, nodes-graph pairs~\cite{velickovic2018deep,peng2020graph,sun2019infograph} or identical node pairs~\cite{zhu2020deep,zhu2020graph}. Different from previous methods, contrastive pairs are identical channel pairs in our work.

{\bf Proximity preserving} is a means for graph learning models to retain and learn from the similarity and relation between nodes. Based on different similarity measures, the models may be classified into three categories: 1) Preserving proximity in the graph structure. Network embedding methods like DeepWalk~\cite{perozzi2014deepwalk}, LINE~\cite{tang2015line}, and node2vec~\cite{grover2016node2vec} preserve this kind of proximity by maximizing the co-occurrence probability of nodes and their neighborhoods. Message passing GNNs~\cite{kipf2017semi,velickovic2018graph,hamilton2017inductive} preserve this proximity by local feature smoothing. 2) Preserving proximity in feature space, where nodes with similar features are in close proximity in the graph's feature space. For this purpose, AM-GCN~\cite{wang2020gcn} and SimP-GCN~\cite{jin2021node} construct a $k$NN graph based on the feature matrix, and then input the generated $k$NN graph together with the original graph to jointly train the GNN model. 3) Preserving proximity in topology space, where nodes with similar local structures are in proximity. GraphWave~\cite{donnat2018learning} leverages the diffusion of spectral graph wavelets to model structural similarity. struc2vec~\cite{ribeiro2017struc2vec} uses a hierarchy to capture structural similarity at different scales. RolX~\cite{henderson2012rolx} aims to recover a soft-clustering of nodes into a specific number of distinct roles using recursive structural feature extraction. In general, proximity in the graph structure is explicitly reflected by the edges in the graph, while feature proximity and local topology proximity are implicit. In this paper, we aim to exploit the three types of proximity simultaneously, under an unsupervised setting.


\section{Preliminaries}\label{sec:pre}
\paragraph{Notation} Let $G = (V, E, \mathbf{X})$ denotes an unweighted attributed graph, $V = \{v_1,\cdots, v_N\}$ is a set of nodes with $|V| = N$, $E \subseteq V \times V$ is a set of edges, and $\mathbf{X} = \{\mathbf{x}_1,\cdots, \mathbf{x}_N\}$ is a set of attribute vectors, where $\mathbf{x}_i \in \mathbb{R}^F$ is a $F$-dimensional vector for node $v_i$. The node feature matrix can also be viewed as a graph signal\cite{shuman2013emerging} with $F$ channels defined on $G$. Usually, $V$ and $E$ can be jointly denoted as an adjacency matrix $A \in \mathbb{R}^{N \times N}$: for undirected graph, if $e_{ij} = (v_i, v_j) \in E$, then $A_{ij} = A_{ji} = 1$, otherwise, $A_{ij} = A_{ji} = 0$; for directed graph, $A_{ij} = 1$ if and only if $v_i$ links to $v_j$ with an outgoing edge. Thus, the graph can also be written as $G = (A,\mathbf{X})$. The sets of outgoing and incoming neighbors of a node $v_i$ are written as $N^{out}_i$ and $N^{in}_i$, respectively. 

\paragraph{Unsupervised graph representation learning (GRL)} Given a positive integer $d$ as the expected embedding dimension ($d \ll N $ usually) and a graph $G$ with one-hot node labels $Y = \{y_1,\cdots, y_N\}$, where $y_i \in \mathbb{R}^C$ and $C$ is the number of labels, under a unsupervised setting, GRL aims to learn a $d$-dimensional representation for each node $v_i \in V$ with all labels unseen. 
 
\paragraph{Graph Neural Networks} Most graph neural networks, as instances of message passing neural networks (MPNN)~\cite{gilmer2017neural}, perform neighborhood aggregation to locally smooth features of each node. In this way, the learned representation vectors can preserve the proximity in the graph structure. \citet{kipf2017semi} add the undirected assumption to explain GNNs in terms of spectral analysis. In fact, MPNN-based GNNs are compatible with both directed and undirected graphs, where undirected edges are treated as bidirectional edges in undirected graphs. Thus, the embedding of $l$-th layer of directed graph neural networks for node $v_i$ can be defined as:
\begin{equation}
	h_{i}^{(l)}=\operatorname{COMB}^{(k)}\left(h_{i}^{(l-1)}, \operatorname { AGG }^{(l)}\left(\left\{\left(h_{i}^{(l-1)}, h_{j}^{(l-1)}, e_{j i}\right): v_j \in N^{in}_i\right\}\right)\right ),
	\label{digcn}
\end{equation}
where $\operatorname{COMB}(\cdot)$ denotes the representation updating function, $\operatorname{AGG}(\cdot)$ denotes the aggregation function. We use $W^{(l)}$ to denote the learnable parameters in the $l$-th layer.




\paragraph{Proximity Graph} We reconstruct the graph in both feature and topology spaces. As shown in \Cref{fig:example}, in the feature space each node only links to the $k_f$ most similar nodes to build a \textbf{feature proximity graph} (FPG), denoted $G_f = (A_f, \mathbf{X})$. In the topology space, each node links to the top $k_t$ nodes with most similar local topology to build a \textbf{topology proximity graph} (TPG), denoted $G_t = (A_t, \mathbf{X})$. As such, $G_f$ and $G_t$ are two views of the input graph $G$. Note that the similarity ranking may be asymmetric so the generated views are directed. In addition, proximity graphs are structural views, and therefore the augmentation is only applied to the structure of the graphs rather than the initial node features.

\section{The Proposed Model}

\begin{figure*}[t]
	\centering
	\includegraphics[width=1.1\linewidth]{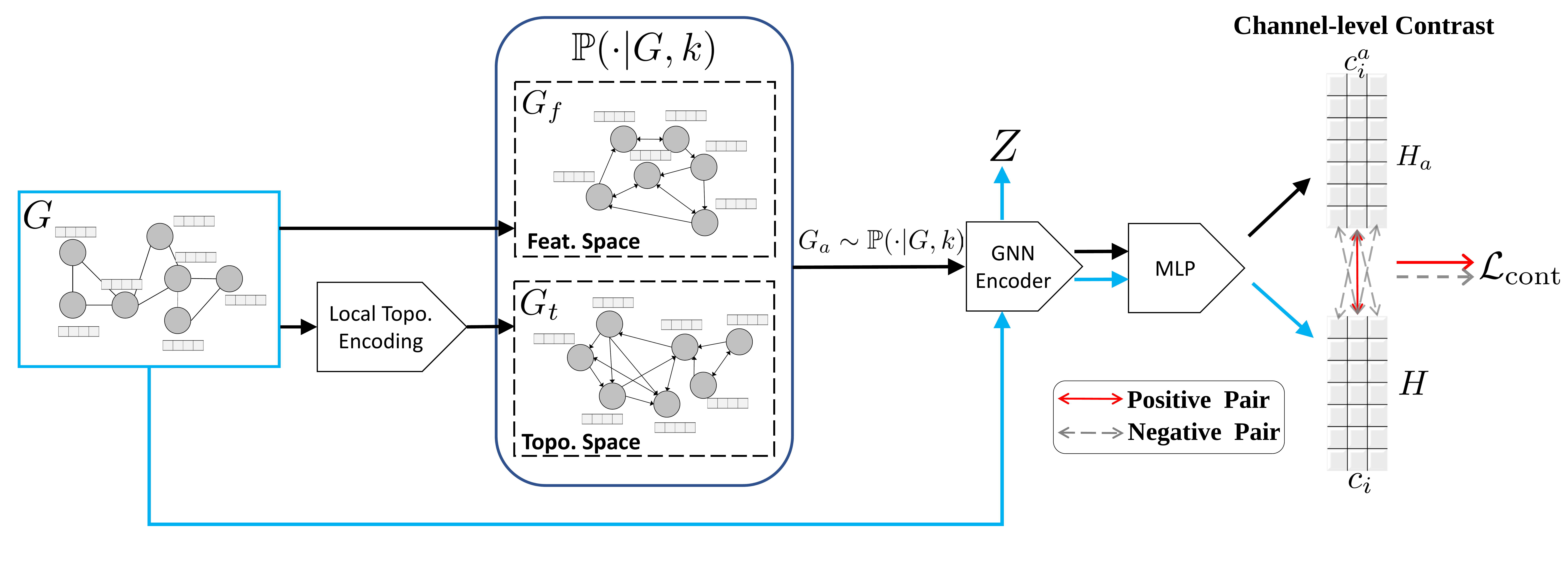}
	\caption{Overview of FT-GCL. The input graph $G$ and its view $G_a$ are fed into the shared GNN encoder and MLP ($G_a$ is FPG $G_f$ or TPG $G_t$). The final contrastive objective is to maximize the agreement of the distribution in the identical channels such as $c_i$ and $c^a_i$, and distinguish different channels.}
	\label{fig:framework}
\end{figure*}

In this section, we propose a simple yet powerful contrastive learning model, called \textsc{Feature \& Topology proximity preserving Graph Contrastive Learning} (FT-GCL). In this framework, as shown in \Cref{fig:framework}, the input graph $G$ and one of its views are embedded by a shared GNN encoder. Then, the contrastive objective aims to enforce the consistency between the representations of the original graph and its views. 
	
\begin{figure}[t]
	\centering
	\begin{minipage}[t]{0.46\textwidth}
		\vspace{0pt}
    	\centering
    	\includegraphics[width=0.7\textwidth]{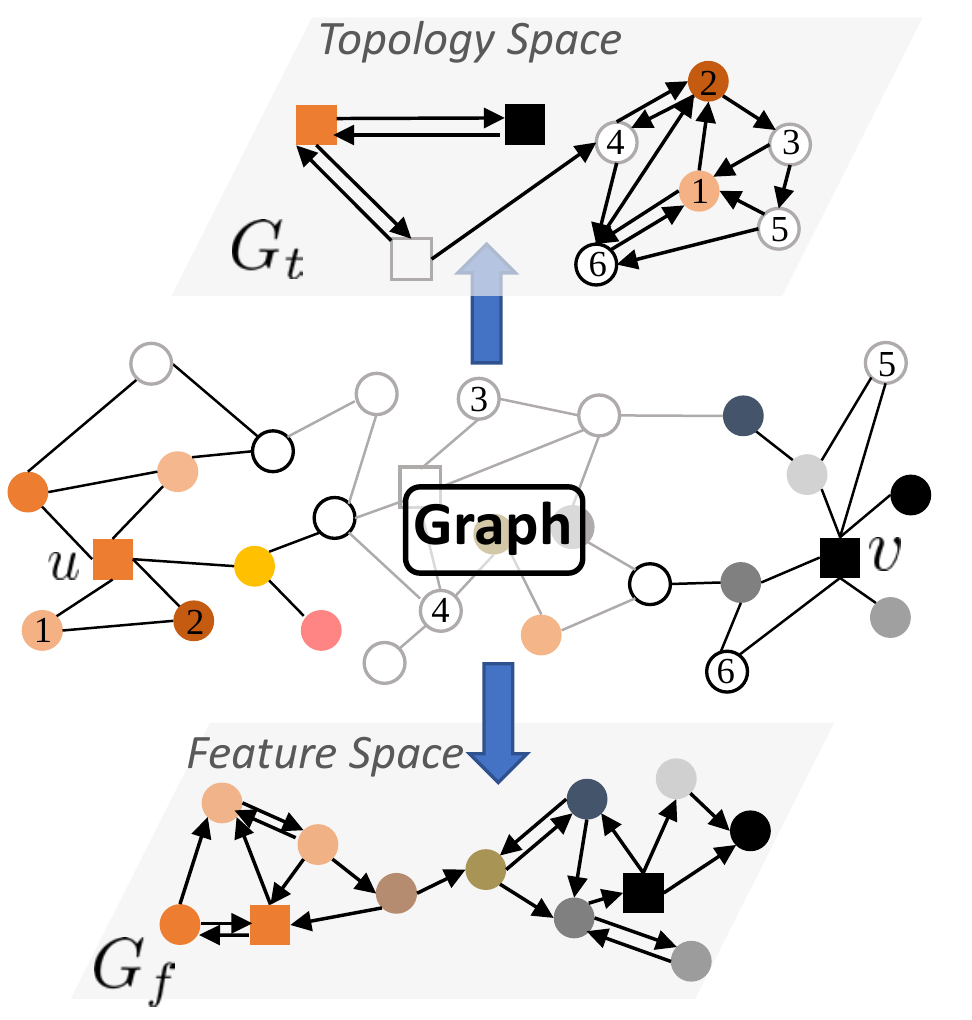}
    	\caption{The rectangle nodes $u$ and $v$ have similar local structures, and nodes with similar color have similar features.}
    	\label{fig:example}
	\end{minipage}
	\hfill
	\begin{minipage}[t]{0.48\textwidth}
		\vspace{0pt}
    	\centering  
    	\subfigure[]{
    		\label{fig:bb_graph}
    		\includegraphics[width=0.46\linewidth]{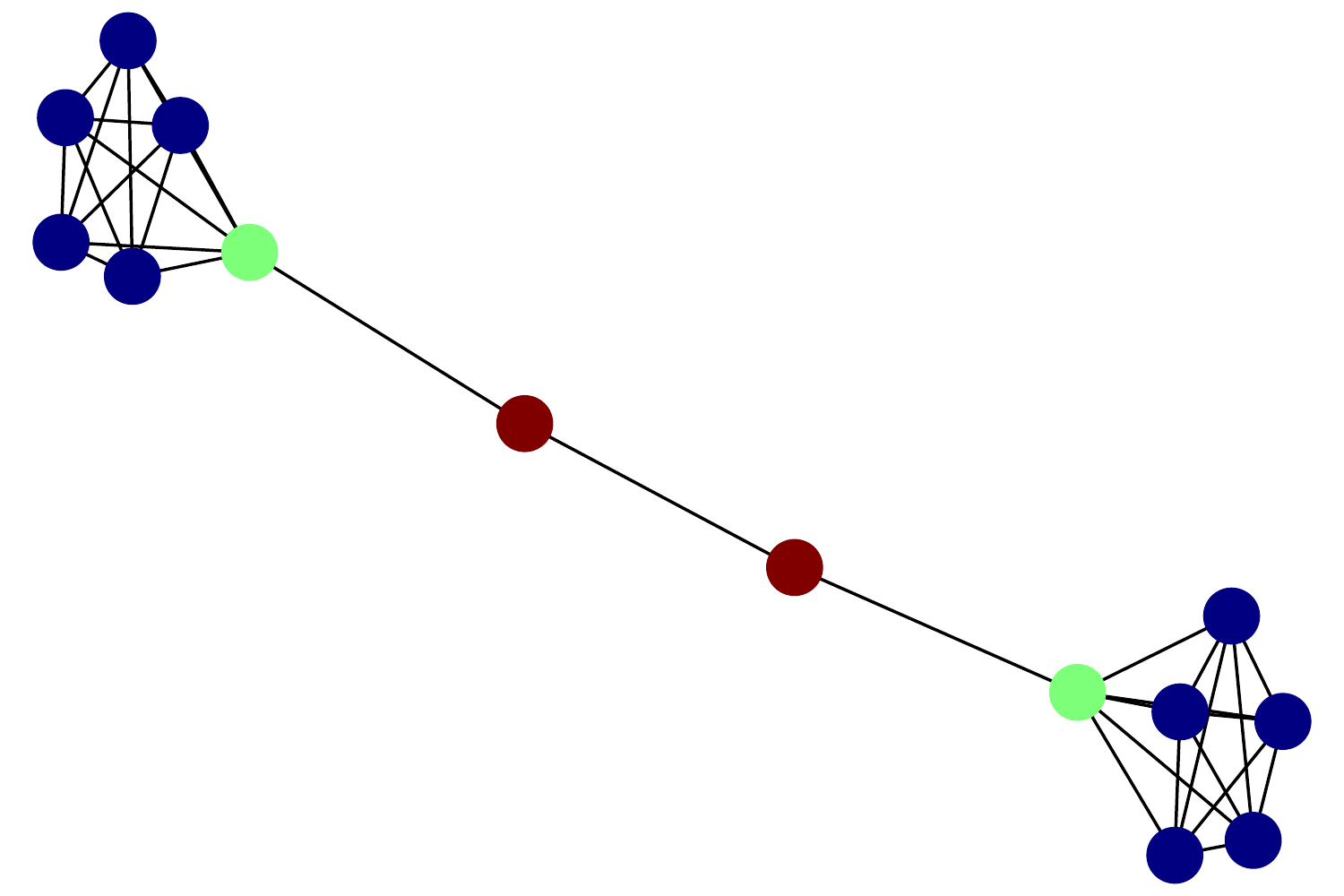}}
    	\subfigure[]{
    		\label{fig:bb_embedings}
    		\includegraphics[width=0.46\linewidth]{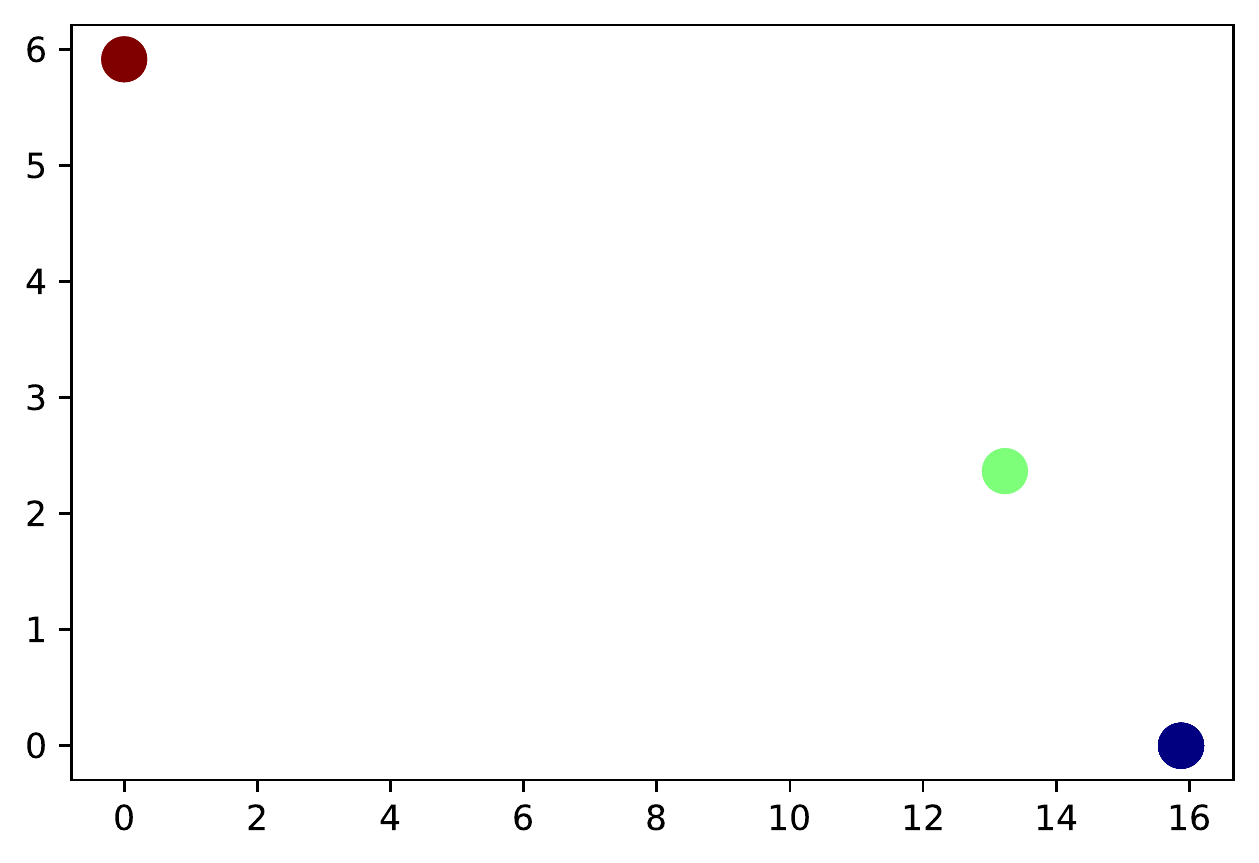}}
    	\caption{(a) Barbell graph $B(6,2)$, where nodes with the same first-order neighbor have the same color (b) Local topology representations in $\mathbb{R}^2$ (topology space of the graph) learned by factorizing the kernel matrix with Nyström approximation. The result shows that nodes with the same local structure coincide well in the topology space.}
    	\label{fig:barbell}
	\end{minipage}
\label{fig:exp_ft}
\end{figure}

\paragraph{Step-1: Views generation} In this stage, we generate FPG and TPG of $G$ as two views to reflect proximity in the latent space. \Cref{fig:example} shows an example where nodes $u$ and $v$ have similar local topology: they have degrees 5 and 6, and are both connected to 2 triangles, therefore they should be connected in the topology space. To that end, we first encode the structural roles of each node as a vector in topology space. Although several structural role embedding methods~\cite{ribeiro2017struc2vec, henderson2012rolx,narayanan2016subgraph2vec} can do the job, they require separate back-propagation or enumeration operations, which are too heavy as a preprocessing step. Here, we provide an alternative method based on graph kernel.

We first extract local subgraphs for all the nodes, where the subgraphs can be $r$-hop egonets (i.e., the induced subgraph surrounding each node) or induced by a set of random walk (RW) sequences start from each node. In effect, different choices do not affect the results significantly and RW is more efficient for dense graphs. Let $\mathcal{S} = \{S_1, S_2, \cdots, S_N\}$ be the set of extracted local subgraphs for each node, we adopt the Weisfeiler-Lehman (WL) subtree kernel~\cite{shervashidze2011weisfeiler} with a fixed number of iterations on the subgraph set $\mathcal{S}$. Finally, we can obtain a symmetric positive semi-definite kernel matrix $K \in \mathbb{R}^{N \times N}$, where $K_{ij}$ denotes the similarity between local subgraphs $S_i$ and $S_j$, which can be seen as the local topological similarity between $v_i$ and $v_j$. Then, we directly factorize $K$ to obtain the representation of each subgraph. This leads to a high computational complexity of $\mathcal{O}(N^3)$. We use Nyström approximation~\cite{williams2001using, nikolentzos2018kernel} to reduce the cost to $\mathcal{O}(m^2N)$, where $m \ll N$. The Nyström method only uses a small subset of $m$ columns (or rows) of the $K$, such that $K \approx RR^\top$ where $R \in \mathbb{R}^{N \times m}$. Finally, each row of $R$ represents the coordinate of the corresponding node in the graph's topology space, and therefore serves as the local topology representation of the node. 

\Cref{fig:barbell} shows an experiment on a barbell graph $B(6,2)$, where the local subgraphs are nodes' 1-hop egonets. Based on $R$, we construct a directed $k$NN graph where each node links to top $k_t$ similar nodes using outgoing edges, denoted $G_t = (A_t, \mathbf{X})$. Note that in the views, $R$ is only used to generate the structure of $G_t$ and node features are still $\mathbf{X}$.

In the feature space, a proximity graph $G_f$ is generated from $G$ based on node features $\mathbf{X}$. Both proximity graphs use cosine similarity as the distance metric. In the views generation stage, the computational cost mainly comes from calculating the vector similarity for the $k$NN graphs, which is $\mathcal{O}((F+m)N^2)$. Fortunately, fast approximiation~\cite{chen2009fast,zhang2013fast} and parallelization~\cite{dong2011efficient} methods have been well studied to construct $k$NN graphs efficiently. For example, we are able to construct $k$NN graphs with local sensitive hashing~\cite{zhang2013fast} in $\mathcal{O}(\ln(\max\{F,m\} + \log N))$.

Two graph views $G_f$ and $G_t$ with $k_f = k_t = k$ are denoted $\mathbb{P}(\cdot | G,k)$. In each training step, we randomly set a $k$ value and generate a view $G_a \sim \mathbb{P}(\cdot | G,k)$, where $G_a$ is alternately set to $G_f$ or $G_t$. The randomness of $k$ helps improve the view diversity, such that proximity with different scales can be preserved. However, repeatedly constructing $k$NN graphs in each training step is expensive. We limit the values of $k_f$ and $k_t$ to $k_{max}$, i.e., $k_f, k_t \leq k_{max}$. We only need to find top-$k_{max}$ neighbors for each node in the feature and the topology spaces before training, which is a single-time effort. Then in each training step, FPG with $k_f$ or TPG with $k_t$ can be easily obtained from the preprocessed $k_{max}$ neighbors by masking a certain number of neighbors at the end of the similarity ranking. In practice, $k_{max}$ is set manually as a hyperparameter and $k_{max} < 10$ is found to work well.



\paragraph{Step-2: Shared GNN encoder} The view $G_a \sim \mathbb{P}(\cdot | G,k)$ together with $G$ are fed into a (Di)GNN encoder $\mathrm{GNN}(A, X)$, where each node aggregates the messages propagated from its in-neighbors. The design of the GNN encoder is flexible. We find that GAT~\cite{velickovic2018graph} or GraphSAGE~\cite{hamilton2017inductive} both perform well. The trainable parameter matrix $W^{(l)}$ in each layer is shared between two graphs during training. 

\paragraph{Step-3: Shared MLP} After the GNN encoder, we use a shared Multi-Layer Perceptron (MLP) $g(\cdot)$ as a projection head~\cite{chen2020simple, you2020graph,hassani2020contrastive} to enhance the expressive power of the output representations.


\paragraph{Step-4: Channel-level contrastive objective}

The shared MLP outputs two representation matrices $H, H_a \in \mathbb{R}^{N \times d}$ for $G$ and its sampled augmented view $G_a$, which can be seen as two signals with $d$ channels on two graph views. Different from previous work~\cite{zhu2020graph, zhu2020deep, jovanovic2021towards} that focuses on node-level contrasting, which suffers from high computation cost, we offer an alternative method that generates contrastive pairs at the channel level. Specifically, as shown in \Cref{fig:framework} the $i$-th channels of two output signals, denoted $c_i$ and $c^a_i$, are the $i$-th columns of $H$ and $H_a$, respectively. Then the contrastive objective aims to maximize the consistency between two representation matrices, such that the distributions of identical channels $c_i$ and $c^a_i$ as inter-graph positive pairs are to be pulled together, and different channels $c_i$ and $c^a_{j,j \neq i}$ as negative pairs to be pushed away. Hence, the loss function of pair-wise channels $(c_i, c^a_i)$ is defined as:
\begin{equation}
	\ell_i=-\left(\log \frac{\exp(\phi(c_i, c^a_i)/ \tau)}{\sum_{j=1, j \neq i}^d  \exp(\phi(c_i, c^a_j)/\tau) } + \log \frac{\exp(\phi(c_i, c^a_i)/ \tau)}{\sum_{j=1, j \neq i}^d \exp(\phi(c_j, c^a_i)/\tau)}\right)
\label{eq:cl_loss}
\end{equation}
where $\tau$ is the temperature hyper-parameter to scale the cosine similarity $\phi(\cdot,\cdot) = \frac{\cdot^\top\cdot}{\lVert \cdot \rVert \lVert \cdot \rVert}$. \citet{chen2020simple} find that an appropriate $\tau$ can help the model learn from hard negatives. Considering all $d$ channels, the total contrastive loss function is as follows,
\begin{equation}
	\mathcal{L}_{cont} = \frac{1}{d} \sum_i^d \ell_i
	\label{eq:total_cl}
\end{equation}


\paragraph{Prediction} The GNN encoder and MLP are all trained in an unsupervised manner. After the training stage, the original graph $G$ is fed into the GNN encoder to generate the resulting node embeddings $Z$, as shown in the \Cref{fig:framework}. Then we can apply $Z$ to downstream prediction tasks such as node classification or link prediction. See \cref{app:algo} for the summarized algorithm.

\subsection{Theoretical Analysis}\label{sec:theor}
We create two graphs $G_f$ and $G_t$ as two views of $G$, and then maximize the agreement between the output embedding matrices of $G$ and a view. By this means, the GNN encoder is able to capture richer semantic information (i.e., feature similarity and structural equivalence), which may boost the subsequent prediction tasks. In other words, minimizing the contrastive objective in Eq. (\ref{eq:total_cl}) aims to maximize the ``correlation'' between the input graph $G$ and its proximity graph $G_{f}$ or $G_t$. However, what does ``correlation'' specifically mean? In the following, we theoretically discuss the connection between the loss function of FT-GCL and InfoNCE~\cite{oord2018representation} in detail to answer this question, and prove that minimizing our proposed channel-level contrastive objective can be viewed as a kind of maximizing a lower bound of the Mutual Information (MI) between $G$ and its views.

Given two view sets $A$ and $B$, both with the same cardinality $n$, there exists a bijection $\pi: A \to B$ such that each view of one set is positively paired with exactly one view of the other set. Any other inter-set pairs are negative. Under this condition, InfoNCE~\cite{oord2018representation, kong2019mutual} which is upper bounded by MI, can be defined as:
\begin{equation}
	I_{NCE} (A;B) = \mathbb{E}_{(a, b) \sim \mathbb{P}(A, B)}\left[f_{\boldsymbol{\theta}}(a, b)-\log \mathbb{E}_{\tilde{b} \sim \mathbb{P}(B)} \exp f_{\boldsymbol{\theta}}(a, \tilde{b})\right] \leq I(A,B)
	\label{eq:mi}
\end{equation}
where $f_\theta (\cdot) \in \mathbb{R}$ is a function parameterized by $\theta$, $\mathbb{P}(A, B)$ is the distribution of positive pairs. Then our contrastive objective Eq. (\ref{eq:total_cl}) can be rewritten as the expectation form:
\begin{equation}
	\begin{aligned}
		-\mathcal{L}_{cont} =& 2\mathbb{E}_{\mathbb{P}(c_i, c^a_i)} \frac{\phi(c_i, c^a_i)}{\tau}-\mathbb{E}_{\mathbb{P}(c_i)}\log \mathbb{E}_{\mathbb{P}(c^a_j)}\exp(\phi(c_i, c^a_j)/\tau)\\
		&-\mathbb{E}_{\mathbb{P}(c^a_i)}\log \mathbb{E}_{\mathbb{P}(c_j)}\exp(\phi(c_j, c^a_i)/\tau) - 2\log d,
	\end{aligned}
	\label{eq:cl_mi}
\end{equation}
where $c^{a} = g(\mathrm{GNN}(G_a))$, dimension $d$ is a constant. $\mathbb{P}(c_i, c^a_i)$ is a discrete uniform distribution of size $d$, so $\mathbb{P}(c_i, c^a_i) = \mathbb{P}(c^a_i) = \mathbb{P}(c_i)$.  Let $f_\theta(c^a,c^b) = \phi(g(\mathrm{GNN}(G_a)),g(\mathrm{GNN}(G_b)))/\tau$, where $\theta$ parametrized by $\tau$ and the MLP $g(\cdot)$. Since there is bijection between column vector sets of $H$ and $H_a$ as positive pairs,  then Eq. (\ref{eq:cl_mi}) can be transform to the InfoNCE form in Eq. (\ref{eq:mi}) as:
\begin{equation}
	\begin{aligned}
		-\mathcal{L}_{cont} =& 2\mathbb{E}_{\mathbb{P}(c_i, c^a_i)} f_{\theta}(c_i, c^a_i)-\mathbb{E}_{\mathbb{P}(c_i)}\log \mathbb{E}_{\mathbb{P}(c^a_j)}f_{\theta}(c_i, c^a_j) -\\
		&\mathbb{E}_{\mathbb{P}(c^a_i)}\log \mathbb{E}_{\mathbb{P}(c_j)}f_{\theta}(c_j, c^a_i) \leq 2I(H, H_a),
	\end{aligned}
	\label{eq:cl_mi2}
\end{equation}
here we omit the constant $2\log d$. The objective function of FT-GCL aims to maximize Eq. (\ref{eq:cl_mi2}), that is equivalent to maximizing a lower bound of the mutual information between the output representation of $G$ and that of its proximity graph $G_a$.

\section{Experiments and Analysis}
\begin{wrapfigure}{r}{0.38\textwidth} 
	\vspace{-30pt}
	\begin{center}
		\includegraphics[width=0.453\textwidth]{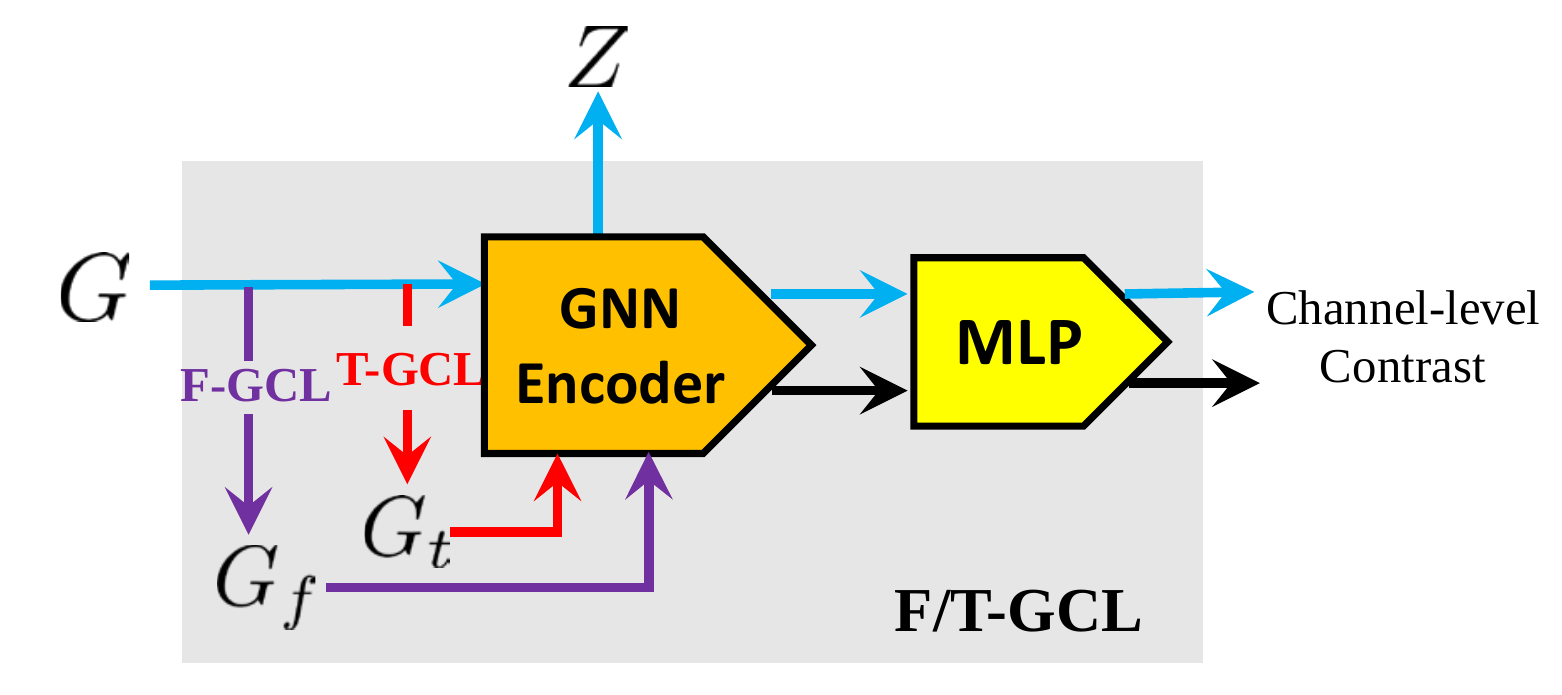}
	\end{center}
	\caption{Framework of F-GCL and T-GCL.}
	\label{fig:ablation}
\end{wrapfigure}

\subsection{Effect of proximity graphs: ablation models}

In FT-GCL, we do contrastive learning between the original graph and a sampled view in each training step, which raises a question: What is the effect of individual proximity graph on learning performance? To answer this question, we propose two ablation models: \textsc{Feature proximity preserving GCL} (F-GCL) and \textsc{Topology proximity preserving GCL} (T-GCL), as shown in \Cref{fig:ablation}. In F-GCL, the original graph only contrasts with the feature proximity graph, while in T-GCL, the original graph only contrasts with the topology proximity graph. From these two models, we examine how feature proximity and topology proximity information benefit graph representation learning.

\subsection{Node classification on assortative graphs}
\subsubsection{Experimentational setup}
\paragraph{Datasets}
We conduct experiments on the following seven real-world assortative graphs which are commonly used by existing work on graph learning. \Cref{tab:datasets} summarizes the statistics of these benchmark datasets. See \cref{app:dataset_ass} for detailed description.

\paragraph{Evaluation protocol}\label{sec:eva_pro}

For WikiCS, we directly use 20 public train/validation splits provided by~\citet{mernyei2020wiki}. For other datasets, we use ten splits of the nodes into 10\%/10\%/80\% for train/validation/test nodes over 10 random seeds, then we run 10 times for each split to report average accuracy (Acc) and macro F1-score (Macro-F1). In the testing stage, the original graph is directly input into the GNN encoder and output a node representation matrix $Z \in \mathbb{R}^{N \times d^\prime}$, as shown in \Cref{fig:framework}, then the embeddings of the training set are used to train a $l_2$-regularized logistic regression classifier and give the results of classification on the test nodes. We provide implementation details in~\Cref{app:imp_det}, with hyper-parameters settings in \Cref{app:hyper}.

\begin{table}[h!]
	\begin{center}
		\caption{Assortative datasets statistics.}
		\label{tab:datasets}
		\scalebox{0.9}{
			\begin{tabular}{ccccccccc} 
				\toprule			
				&\textbf{Cora} & \textbf{CiteSeer} & \textbf{PubMed} & \textbf{DBLP} & \textbf{WikiCS} & \textbf{BlogCatalog} & \textbf{ACM} & \textbf{CoauthorCS}\\
				\midrule
				$|V|$&2,708 & 3,327 & 19,717 & 17,716 & 11,701 &5,196 &3,025 & 18,333\\
				$|E|$&5,429 & 4,732 & 44,338 & 52,867 &216,123 &171,743&13,128 & 81,894\\
				$F$  &1,433 & 3,703 & 500    & 1639   & 300    & 8,189 &1,870 &6,805\\
				Classes&7 &6 & 3  &  4 & 10  & 6 & 3 & 15\\
				\bottomrule
		\end{tabular}}
	\end{center}
\end{table}

\subsubsection{Baselines and results}\label{sec:baselines}

We compare FT-GCL with two state-of-the-art unsupervised methods: (1) edge/path sampling based methods, including DeepWalk~\cite{perozzi2014deepwalk}, node2vec~\cite{grover2016node2vec} and LINE~\cite{tang2015line}, and VERSE~\cite{tsitsulin2018verse}, (2) GNN based methods, including GAE/VGAE~\cite{2016variational}, ARVGA~\cite{pan2018adversarially}, MVGRL~\cite{hassani2020contrastive}, DGI~\cite{velickovic2018deep}, GRACE~\cite{zhu2020deep}, and GMI-Adaptive\cite{peng2020graph}.  We also compare FT-GCL with several semi-supervised methods, including GCN~\cite{kipf2017semi}, GAT~\cite{velickovic2018graph}, SGC~\cite{wu2019simplifying} and JK-Net~\cite{xu2018representation}. Limited by space, we report the results of Planetoid datasets in ~\Cref{tab:cora}, which are the most commonly used benchmarks in the literature, and the results of other four datasets can be found in \Cref{app:more_exp}. 

\begin{table}[h!]
	\centering
	\caption{Average classification accuracy (\%) and macro-F1 score (\%) with standard deviation. The training data for each method is shown in \textbf{Input} column. For Planetoid datasets, we run 10 times over ten different random splits in order to comprehensively evaluate the generalization ability of the model on datasets. For clarity, and \crule[green]{0.3cm}{0.3cm} is the champion,  \crule[red!30!white!100]{0.3cm}{0.3cm} is the runner-up.}
	\label{tab:cora}
	\scalebox{0.63}{
		\begin{tabular}{c|c|c|cc|cc|ccccc} 
			\toprule			
			&\multirow{2}{*}{\textbf{Input}}							&\multirow{2}{*}{\textbf{Algo.}}	&\multicolumn{2}{c|}{\textbf{Cora}} & \multicolumn{2}{c|}{\textbf{CiteSeer}} & \multicolumn{2}{c}{\textbf{PubMed}} \\
			&										&					&Acc&Macro-F1						&Acc&Macro-F1					&Acc&Macro-F1\\
			\midrule
			\multirow{14}{*}{\textbf{Unsupervised}}		&\multirow{1}{*}{$X$} 					&Row features 		& $64.23(\pm0.84)$&$60.14(\pm1.25)$  & $64.37(\pm0.75)$& $60.05(\pm1.57)$ & $84.75(\pm0.27)$& $84.67(\pm0.24)$\\  \cmidrule(l){2-9}
			&\multirow{4}{*}{$A$} 					&DeepWalk 			& $74.49(\pm0.33)$&$73.06(\pm0.20)$  &$49.29(\pm1.23)$&$46.86(\pm1.52)$  & $80.25(\pm0.31)$&$78.82(\pm0.33)$\\ 
			&    									&node2vec 			& $75.72(\pm0.26)$&$74.15(\pm0.17)$ & $51.33(\pm1.36)$&$47.37(\pm1.60)$  & $79.88(\pm0.20)$&$78.34(\pm0.24)$\\ 
			&    									&LINE     			& $56.33(\pm0.20)$&$52.11(\pm0.35)$ &$34.74(\pm0.85)$&$31.65(\pm0.93)$ &  $69.31(\pm0.13)$&$66.41(\pm0.32)$\\  
			&										&VERSE				&$75.87(\pm0.70)$&$74.56(\pm1.15)$&$53.37(\pm1.05)$& $49.10(\pm1.03)$&$80.45(\pm0.24)$&$78.93(\pm0.26)$\\\cmidrule(l){2-9}
			&\multirow{10}{*}{$A$,$X$} 				&GAE	      	& $81.37(\pm1.06)$&$79.11(\pm1.26)$ &$68.42(\pm0.56)$& $59.74(\pm0.68)$ &$80.81(\pm0.35)$&$80.10(\pm0.34)$\\
			&										&VGAE			&$80.80(\pm1.30)$&$78.52(\pm2.43)$ &$69.60(\pm0.55)$& $62.73(\pm1.60)$&$82.25(\pm0.28)$&$81.83(\pm0.31)$\\
			&										&ARVGA				&$61.87(\pm2.64)$&$56.05(\pm4.91)$ &$54.17(\pm1.61)$& $46.32(\pm1.77)$&$80.30(\pm0.15)$&$79.79(\pm0.17)$   \\
			&										&MVGRL     			&$83.77(\pm0.67)$&$82.56(\pm0.72)$ &\cellcolor{green}$\mathbf{73.35}(\pm0.47)$&\cellcolor{green} $\mathbf{66.89}(\pm1.24)$ &$85.33(\pm0.16)$&$85.18(\pm0.17)$\\
			&										&DGI      			& $83.61(\pm0.70)$&$82.27(\pm0.85)$ &$70.24(\pm1.44)$ &$65.60(\pm0.97)$&$85.31(\pm0.20)$&$84.74(\pm0.20)$\\
			&										&GRACE    			& $83.24(\pm0.79)$&$81.26(\pm0.51)$ &$71.50(\pm0.79)$& $64.99(\pm1.78)$ & $86.07(\pm0.23)$&$84.79(\pm0.22)$\\
			&										&GMI-Adaptive		& $83.51(\pm0.66)$&$82.39(\pm0.90)$& $72.18(\pm0.35)$&$65.10(\pm0.39)$& $82.97(\pm1.01)$&$79.55(\pm0.39)$\\ \cmidrule(l){3-9}
			&										&\textbf{F-GCL}		&\cellcolor{red!30!white!100}$83.85(\pm0.83)$&$81.85(\pm1.04)$&$ 72.16(\pm0.65)$&$66.44(\pm0.80\textbf{})$&$86.15(\pm0.33)$&\cellcolor{red!30!white!100}$85.59(\pm0.35)$ \\
			&										&\textbf{T-GCL}     &$83.61(\pm0.80)$&\cellcolor{red!30!white!100}$82.65(\pm0.92)$ &$70.76(\pm0.89)$&$65.49(\pm1.12)$&$85.50(\pm0.32)$&$84.94(\pm0.33)$\\
			&										&\textbf{FT-GCL}	&\cellcolor{green}$\mathbf{84.07}(\pm0.58)$&\cellcolor{green}$\mathbf{82.95}(\pm0.79)$& \cellcolor{red!30!white!100}$72.52(\pm1.17)$&\cellcolor{red!30!white!100}$66.79(\pm1.74)$&\cellcolor{red!30!white!100}$86.23(\pm0.59)$& $85.31(\pm0.28)$\\
			\midrule
			\multirow{4}{*}{\textbf{Semi-supervised}}	&\multirow{4}{*}{$A$,$X$,$Y_{train}$}  	&GCN														   			   &$82.57(\pm0.22)$&$81.62(\pm0.16)$&$71.81(\pm0.65)$&$65.10(\pm1.67)$&\cellcolor{green}$\mathbf{86.36}(\pm0.15)$ &\cellcolor{green}$\mathbf{85.77}(\pm0.15)$\\ 
				&									  	&GAT 				& $83.38(\pm0.18)$&$82.09(\pm0.36)$&$72.50(\pm0.70)$&$65.23(\pm0.97)$&$84.70(\pm0.01)$&$84.03(\pm0.38)$\\
				&									  	&SGC 				&$80.28(\pm0.20)$&$78.96(\pm0.28)$&$71.97(\pm0.01)$&$66.31(\pm0.09)$&$81.35(\pm0.19)$&$80.14(\pm0.50)$\\
				&									  	&JK-Net			&$80.10(\pm1.10)$&$79.73(\pm0.81)$&$67.38(\pm1.70)$& $61.95(\pm0.20)$&$82.84(\pm0.50)$&$78.33(\pm0.02)$\\	
			\bottomrule
		\end{tabular}
		}
\end{table}

As shown in~\Cref{tab:cora}, FT-GCL performs the best among unsupervised methods on Cora and PubMed. For the CiteSeer dataset, FT-GCL and MVGRL have comparable performance. Moreover, although without labels to guide the training, FT-GCL can also outperform the classical semi-supervised methods like GCN and GAT, and is only slightly weaker than GCN on PubMed. Note that MVGRL, DGI, GRACE, GMI, GRACE and our FT-GCL are contrastive learning (CL) based GNN models. Traditional network embedding methods like DeepWalk, node2vec and LINE can also be seen as a kind of CL-based method, where positive pairs are nodes with their contexts and negative pairs are nodes with their sampled negative nodes. However, these methods do not achieve desirable results. We think there are two reasons for this: first, node features are not involved in training; second, although they use negative sampling to accelerate the training process, the number of negative samples is usually very small.

The results of the two ablation models are also comparable with the state-of-the-art methods. For Cora, T-GCL achieves the second-best Macro-F1 score. For Pubmed, F-GCL achieves the second-best accuracy and the best Macro-F1 score among all unsupervised methods. The ablation study indicates that preserving either feature proximity or topology proximity during training can improve the quality of the representation vectors.  

\begin{figure}[!tbp]
	\centering
	\begin{minipage}[t]{0.35\textwidth}
		\vspace{0pt}
		\centering
		\includegraphics[trim={0 0.2cm 0 0.2cm}, clip, width=0.9\linewidth]{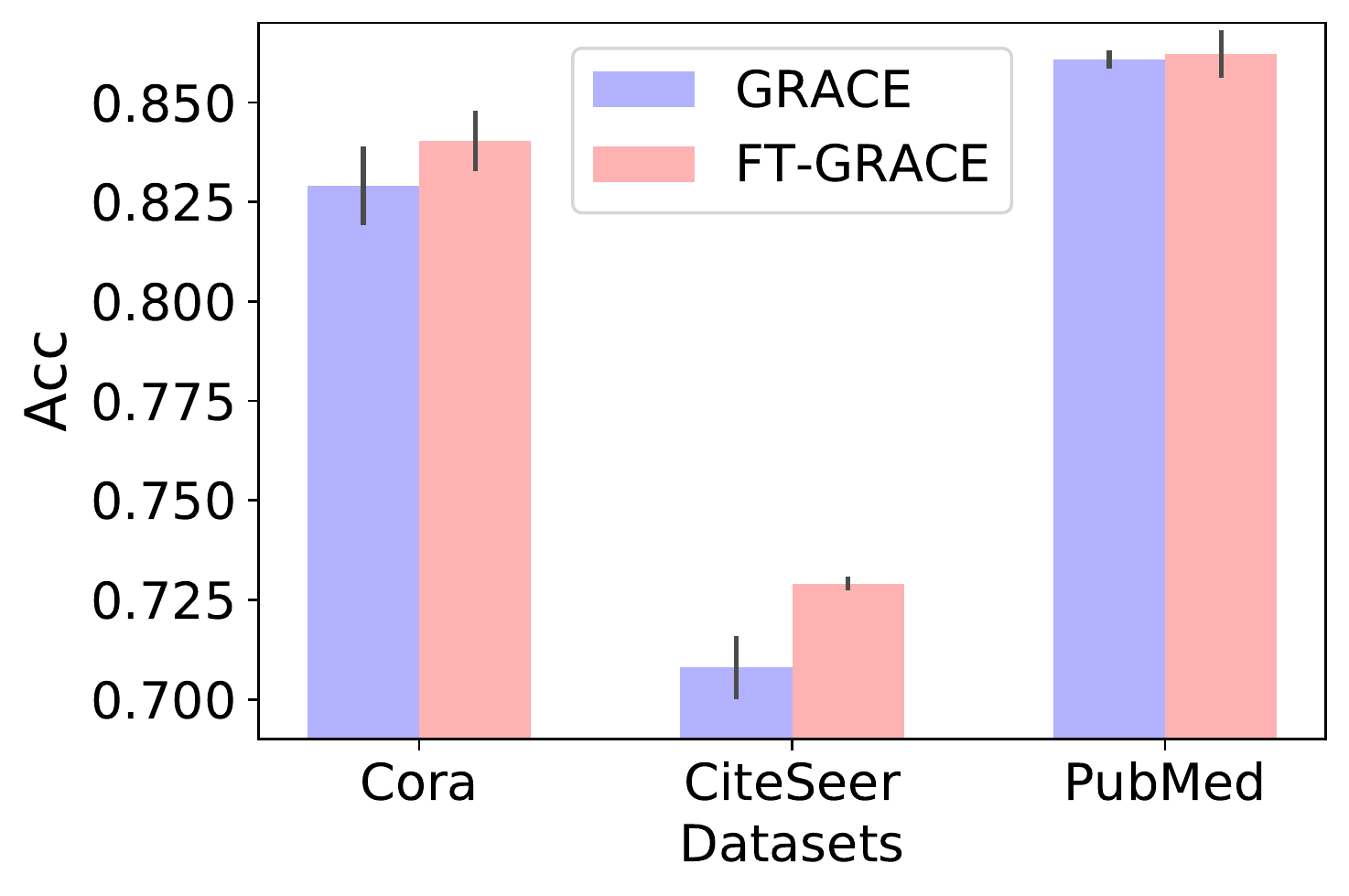}
		\caption{Comparing different graph augmentation schemes on GRACE. }
		\label{fig:ft_grace}
	\end{minipage}
	\hfill
	\begin{minipage}[t]{0.62\textwidth}
		\vspace{0pt}
		\centering
		\includegraphics[width=0.48\linewidth]{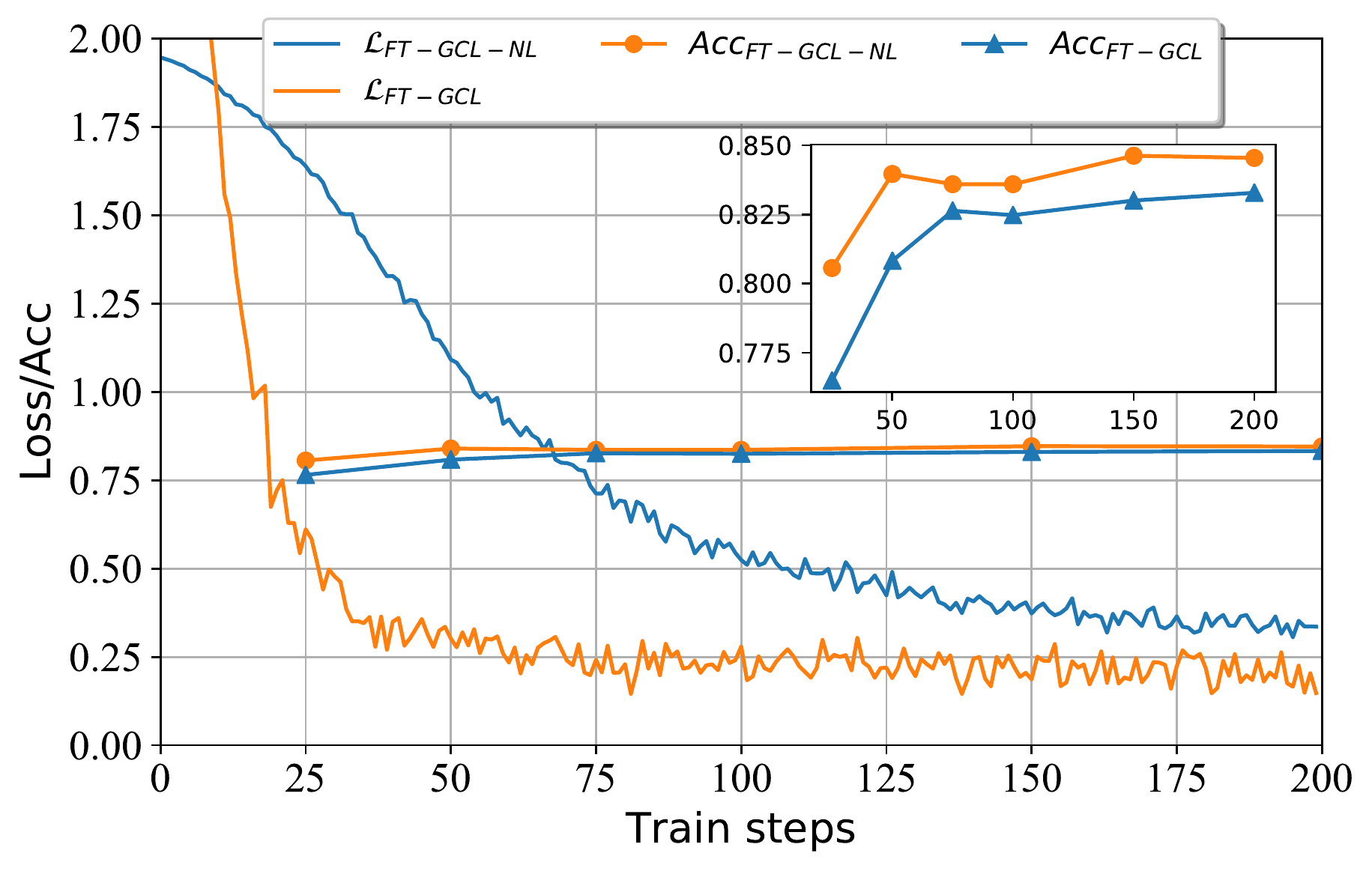}
		\includegraphics[width=0.48\linewidth]{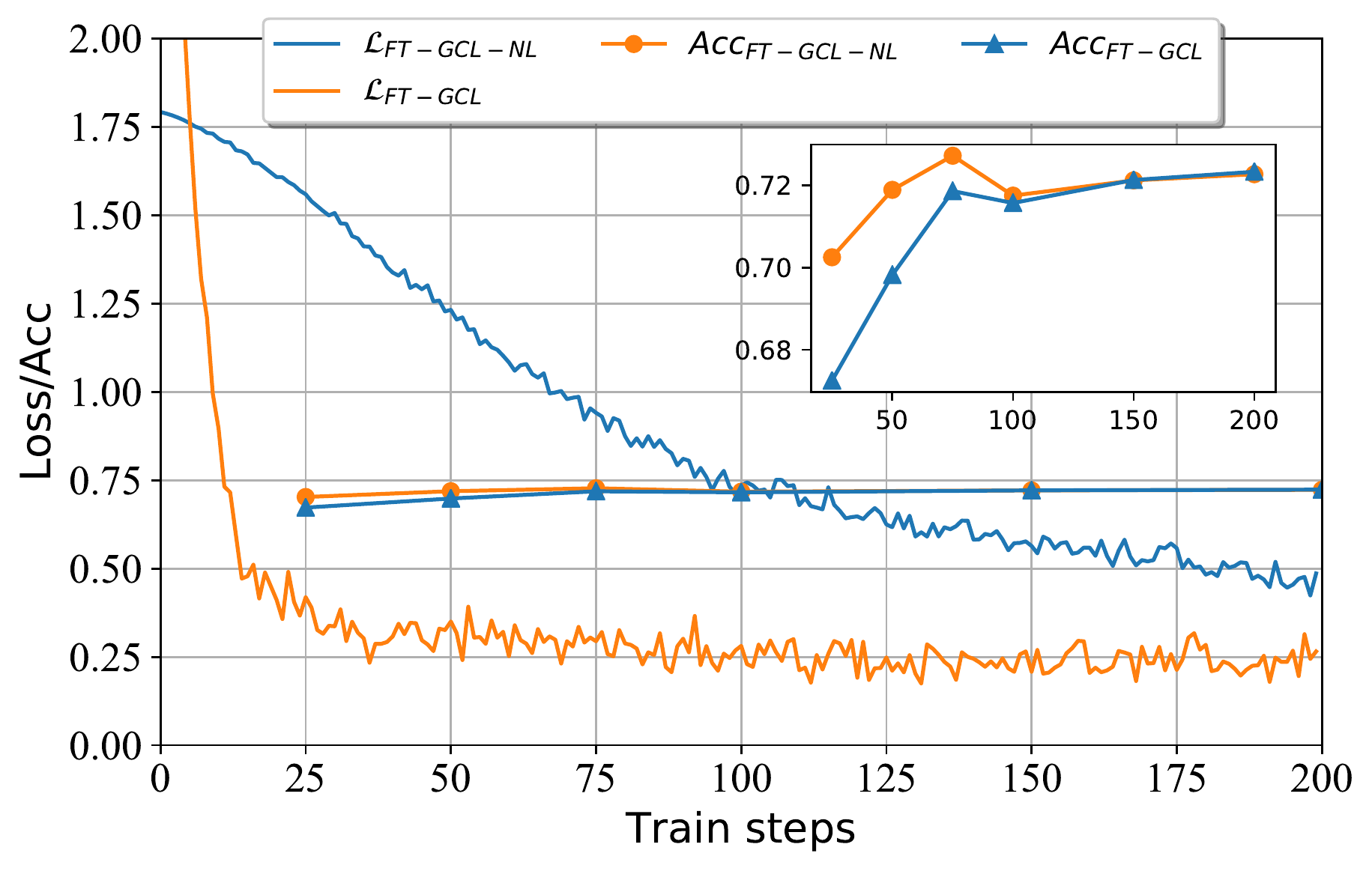}
		\caption{Comparing node-level contrast with channel-level contrast on Cora(left) and CiteSeer(right). $\mathcal{L}_{FT-GCL}$ and $\mathcal{L}_{FT-GCL-NL}$ are loss values of FT-GCL and FT-GCL-NL, respectively. $Acc_{FT-GCL}$ and $Acc_{FT-GCL-NL}$ are their accuracy at several training epochs.}
	\label{fig:nlcont}
	\end{minipage}
\label{fig:exp_ft}
\end{figure}

Given the results of all seven datasets in in~\Cref{tab:cora} and~\Cref{tab:more_exp}, we consider that the strong performance of FT-GCL mainly comes from the idea of node proximity preserving. Compared with GRACE, FT-GCL is different in two respects: graph views generation and contrastive pair construction. Specifically, GRACE applies random corruption on the graph and node features to augment the graph, while FT-GCL uses two proximity graphs and do not modify the original features. In addition, GRACE builds contrastive pairs between two augmented graphs at an inter-graph and intra-graph node level, while FT-GCL only builds contrastive pairs between the original graph and one of the generated graph views at an inter-graph channel level. 

We further replace the graph augmentation scheme of GRACE with proximity graphs as used in FT-GCL, called FT-GRACE. In addition, we replace the contrast scheme of FT-GCL with node-level contrast as used in GRACE, called FT-GCL-NL. Then, we compare the accuracy of GRACE and FT-GRACE to verify the proximity graphs' adaptability, see \Cref{fig:ft_grace}. The results on the three datasets show that proximity graphs also benefit other CL-based models and performs better than random corruption. \Cref{fig:nlcont} shows that FT-GCL achieves better training performance than FT-GCL-NL, according to the speed of convergence. Besides, because FT-GCL-NL performs node-level contrast, the number of negative pairs is $O(N^2)$, where $N$ is the number of nodes. In FT-GCL, the number of negative pairs is reduced to $O(d^2)$, where $d$ is the number of channels of the output features. Usually, $d << N$, which means channel-level contrast is much more efficient than node-level contrast in large graphs.

\subsection{Node classification on disassortative graphs} \label{sec:disassortative}

We also evaluate FT-GCL on a selection of disassortative graphs~\cite{pei2019geom}, namely Texas, Cornell, Wisconsin, and Actor. The statistics of datasets are shown in \cref{tab:dis_datasets}, and more detailed descriptions are shown in \cref{app:dataset_disass}.
For all benchmarks, we use the same setting as~\cite{pei2019geom} (10 public splits with 48\%/32\%20\% of nodes per class for train/validation/test). Other parts of the evaluation protocol follow \Cref{sec:eva_pro}. \Cref{tab:hetero} reports the average accuracy of our methods and other GNN-based models. We also consider a powerful variant of Gemo-GCN~\cite{pei2019geom} which is shown to be powerful on disassortative graphs. The detailed analysis of this experiment is in \Cref{app:more_ana}. 

\begin{table}[htbp]
\centering
	\caption{Disassortative datasets statistics.}
	\label{tab:dis_datasets}
		\begin{tabular}{ccccccccc} 
			\toprule			
				&\textbf{Texas} & \textbf{Cornell} & \textbf{Wisconsin} & \textbf{Actor} \\
				\midrule
				$|V|$			& 183 		& 183 		& 251 		& 7,600				\\
				$|E|$			& 295 		& 280 		& 466 		& 26,752 			\\
				$F$  			& 1,703     & 1,703 	& 1703    	& 932   			\\
				Classes			& 5 		& 5 		& 5  		& 5 				\\
				\bottomrule
		\end{tabular}
\end{table}
\begin{table}[h!]
	\centering
	\caption{Node classification accuracy (\%) with standard deviation on disassortative graphs. \crule[green]{0.3cm}{0.3cm} is the champion, and \crule[red!30!white!100]{0.3cm}{0.3cm} is the runner-up.}
	\label{tab:hetero}
	\scalebox{0.9}{
	\begin{tabular}{c|c|ccccccc} 
		\toprule			
		&\textbf{Algo.}	&\textbf{Texas} & \textbf{Cornell} &\textbf{Wisconsin} & \textbf{Actor} \\
		\midrule
		\multirow{10}{*}{\textbf{Unsupervised}}	
		&GAE	     					&$56.22(\pm4.32)$ &$55.95(\pm5.80)$ &$47.25(\pm6.35)$ & $27.04(\pm1.24)$ \\
		&VGAE							&$54.86(\pm6.17)$ &$55.14(\pm6.42)$ &$48.63(\pm5.46)$ & $27.35(\pm0.99)$ \\
		&ARVGA							&$26.97(\pm1.31)$ &$57.57(\pm4.99)$ &$50.78(\pm6.76)$ & $25.27(\pm1.16)$ \\
		&MVGRL     						&$63.84(\pm4.09)$ &$61.19(\pm3.30)$ &$64.12(\pm7.44)$ & $30.52(\pm0.78)$ \\
		&DGI      						&$57.03(\pm6.99)$ &$55.68(\pm4.86)$	&$54.71(\pm5.85)$ &	$26.80(\pm0.88)$ \\
		&GRACE    						&$56.76(\pm4.83)$ &$53.78(\pm7.20)$	&$48.43(\pm2.16)$ & $27.06(\pm1.19)$ \\
		&GMI-Adaptive					&$53.24(\pm7.74)$ &$61.35(\pm6.05)$ &$53.92(\pm7.19)$ & $29.51(\pm0.82)$ \\ \cmidrule(l){2-6}
		&\textbf{F-GCL}					&$68.65(\pm4.86)$ &\cellcolor{red!30!white!100}$68.46(\pm7.55)$ &\cellcolor{green}$78.82(\pm4.45)$ &	\cellcolor{green}$36.26(\pm0.78)$ \\
		&\textbf{T-GCL}     			&\cellcolor{green}$72.70(\pm4.59)$ &$64.32(\pm7.43)$	&$75.29(\pm4.57)$ & \cellcolor{red!30!white!100}$35.64(\pm0.93)$ \\
		&\textbf{FT-GCL}				&\cellcolor{red!30!white!100}$71.35(\pm5.30)$ &\cellcolor{green}$68.65(\pm7.09)$ &\cellcolor{red!30!white!100}$75.88(\pm4.02)$ & $35.47(\pm0.84)$ \\
		\midrule
		\multirow{5}{*}{\textbf{Semi-supervised}}  
		&Geom-GCN-p						&$67.57$		  &$60.81$          &$64.12$		  & $31.63$\\
		&GCN							&$61.33(\pm7.76)$ &$56.78(\pm4.58)$	&$59.69(\pm7.33)$ &	$30.28(\pm0.69)$ \\ 
		&GAT 							&$58.93(\pm4.37)$ &$59.21(\pm2.34)$ &$53.03(\pm8.37)$ & $26.30(\pm1.32)$\\
		&JK-Net							&$67.49(\pm6.32)$ &$67.93(\pm7.46)$ &$74.52(\pm5.63)$ & $35.10(\pm0.85)$\\
		
		\bottomrule
	\end{tabular}
}
\end{table}

\subsection{Visiualization}
\begin{figure}[htbp]
	\centering
	\subfigure[DeepWalk]{
		\begin{minipage}[t]{0.2\linewidth}
			\centering
			\includegraphics[width=1in]{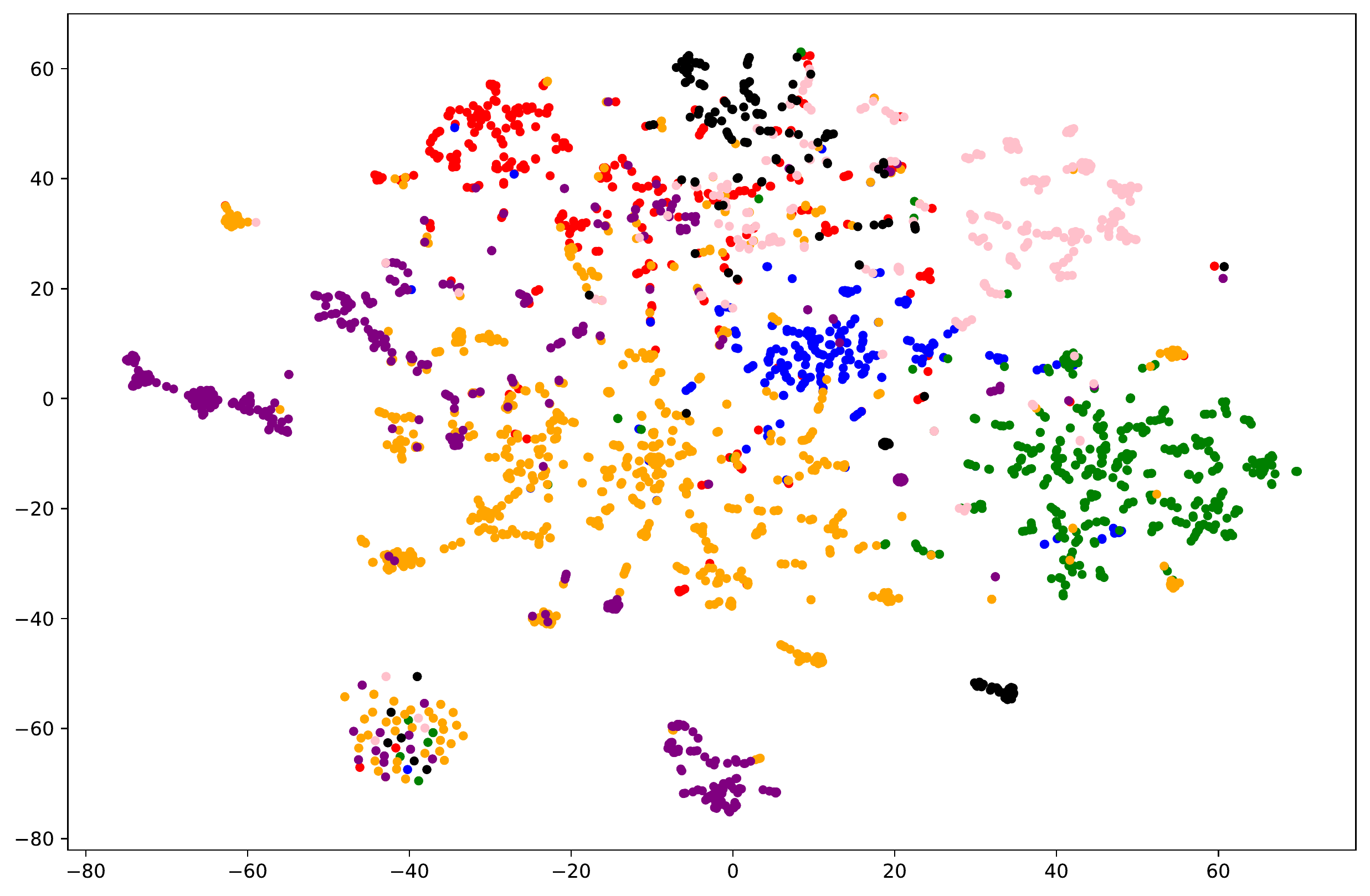}
		\end{minipage}%
	}%
	\subfigure[GAE]{
		\begin{minipage}[t]{0.2\linewidth}
			\centering
			\includegraphics[width=1in]{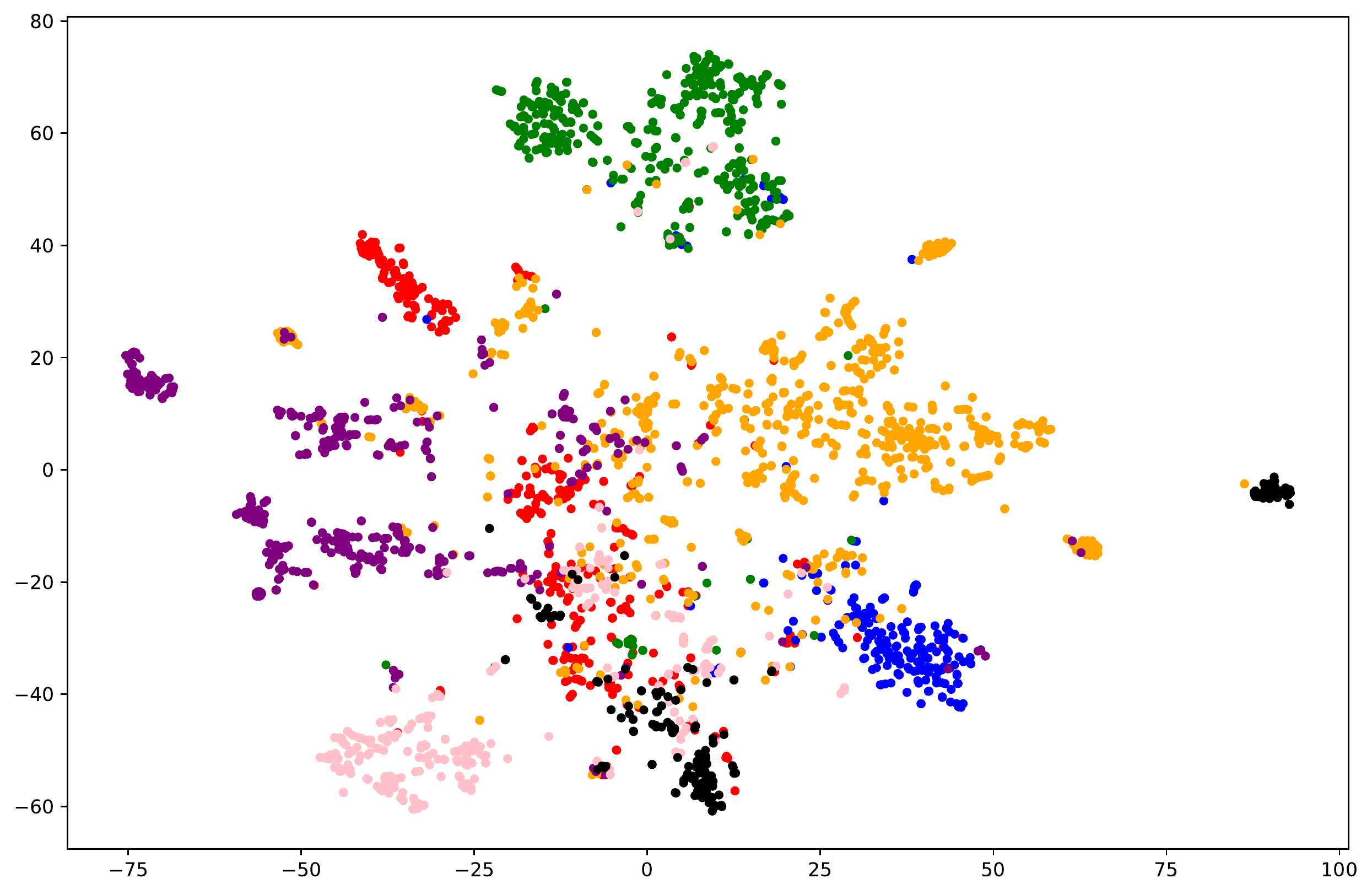}
		\end{minipage}%
	}%
	\subfigure[DGI]{
		\begin{minipage}[t]{0.2\linewidth}
			\centering
			\includegraphics[width=1in]{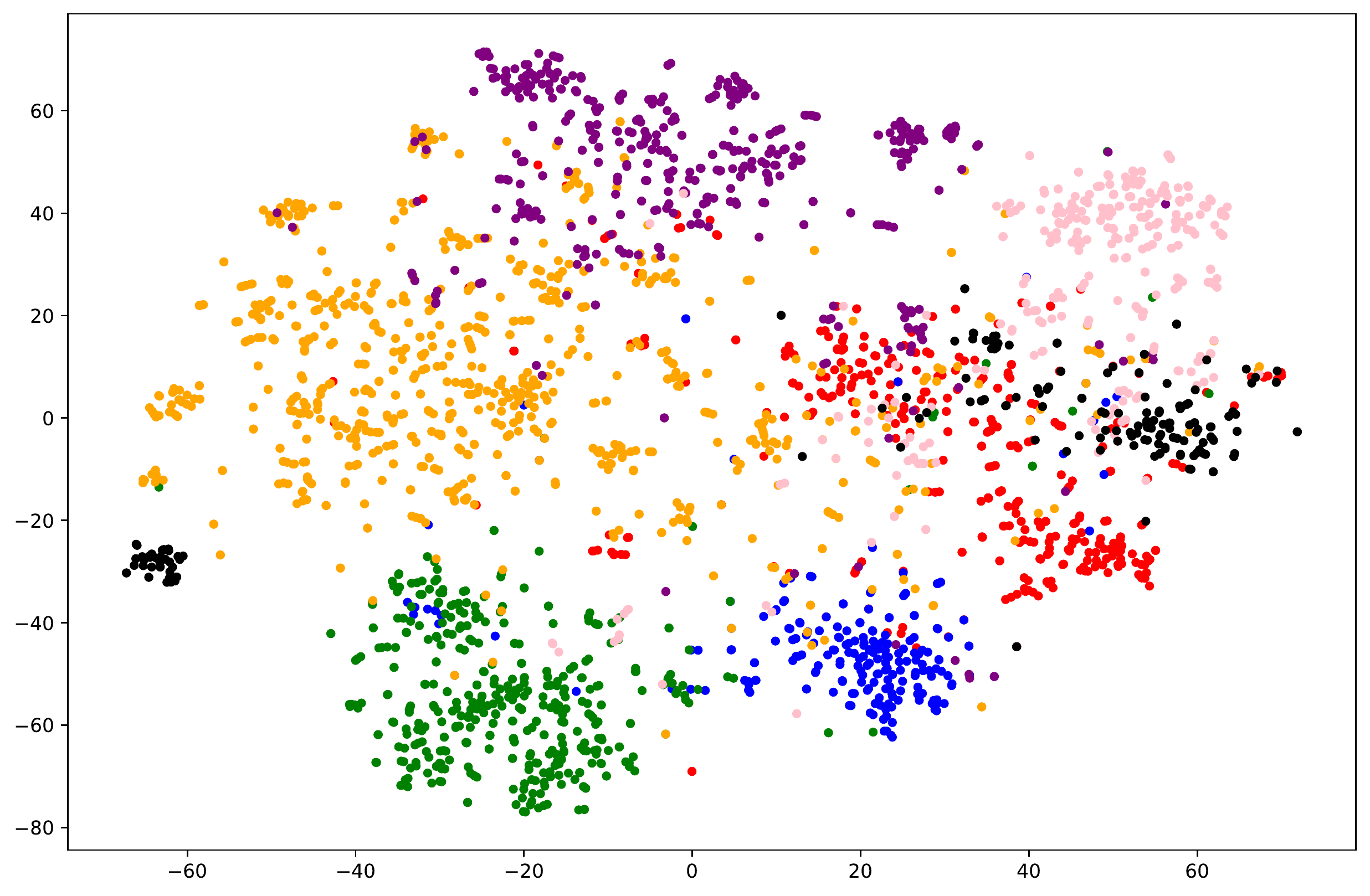}
		\end{minipage}
	}%
	\subfigure[GRACE]{
		\begin{minipage}[t]{0.2\linewidth}
			\centering
			\includegraphics[width=1in]{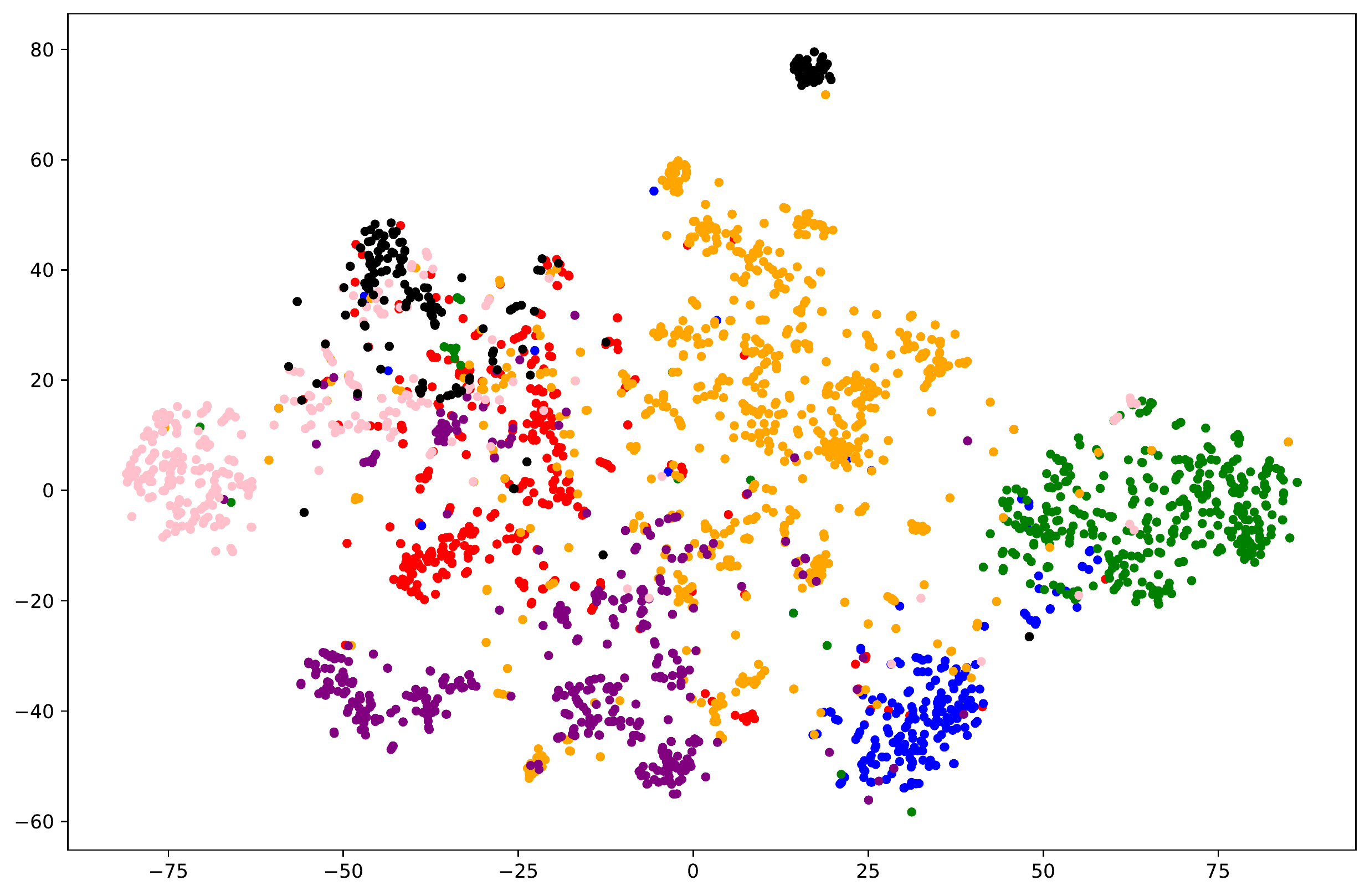}
		\end{minipage}
	}%
	\centering
		\subfigure[FT-GCL]{
		\begin{minipage}[t]{0.2\linewidth}
			\centering
			\includegraphics[width=1in]{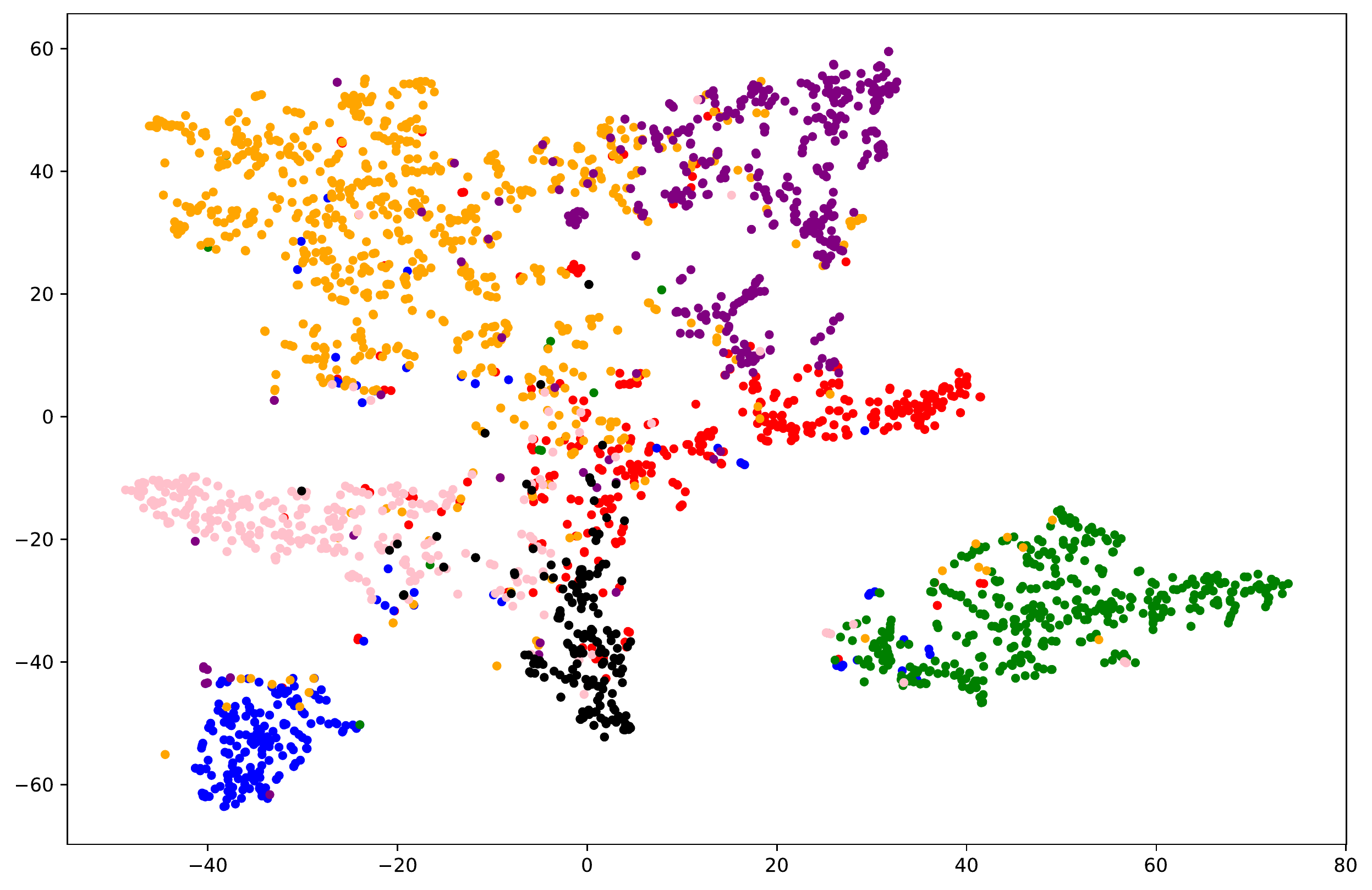}
		\end{minipage}
	}%
	\centering
	\caption{T-SNE visualization of the output embeddings for the Cora dataset.}
	\label{fig:tsne}
\end{figure}

We visualize the output embeddings learned by five unsupervised models using T-SNE~\cite{van2008visualizing}. High-quality embedding vectors exhibit low coupling and high cohesion. The results of Cora in \Cref{fig:tsne} are colored by real labels. From \Cref{fig:tsne}, we find that although GAE and GRACE have relatively high cohesion, the intersection of multiple clusters is more chaotic. FT-GCL performs the best, where the learned representation vectors have higher intra-class similarity and more clear separation among different classes.
\section{Conclusion}\label{sec:conclusion}
In this work, we propose FT-GCL, a self-supervised representation learning framework based on contrastive learning. FT-GCL aims to preserve multiple types of proximity information based on different definitions of node similarity. In FT-GCL, we propose proximity-based graph views based on feature similarity and local topology similarity, and then use a contrastive objective to maximize the agreement of the views and the original graph. In particular, we construct contrastive pairs on a channel level, by which we can greatly reduce the number of negative pairs. We show the connection between InfoNCE and the objective of FT-GCL, which justifies the rationality of our model. Experimental results on 11 datasets show that FT-GCL and its variants are all competitive against state-of-the-art methods. 
\begin{ack}

\end{ack}

\bibliography{references}
\bibliographystyle{biblio}


%

%


\clearpage
\onecolumn
\appendix
\section{Feature and topology proximity graph generation algorithms}\label{app:algo}

\begin{algorithm}[htbp]
	\caption{TPG Generation}
	\label{alg:tpg}
	\begin{algorithmic}[1]
		\State {\bfseries Input:} Graph $G = (V, E, \mathbf{X})$; WL kernel $k$ with $t$ iterations; Number of neighbors $k_t$; Number of walks $\gamma$, Walk length $l$; Number of basis vectors $m$
		\State {\bfseries Output:} Topology proximity graph $G_t = (A_t,\mathbf{X})$
		\For{each $v_i \in V$} 
			\State  $\mathcal{S}_i \gets \mathrm{Subgraph}(\mathrm{RW}(v_i,l, \gamma))$ \Comment{Subgraph induced by random walk node sequences}
		\EndFor
		
	\State $\tilde{\mathcal{S}} \sim \mathbb{P}(\mathcal{S})$       \Comment{Random sample $m$ subgraphs as basis, $|\tilde{\mathcal{S}}| = m$}
	\State $\mathcal{K} \gets k(\mathcal{S},\tilde{\mathcal{S}})$	            \Comment{Compute kernel matrix between $\mathcal{S}$ and $\tilde{\mathcal{S}}$, where $\mathcal{K} \in \mathbb{R}^{N \times m}$}
	
	\State $\tilde{\mathcal{K}} \gets k(\tilde{\mathcal{S}},\tilde{\mathcal{S}})$  \Comment{Comput kernel matrix among basis subgraphs, where $\tilde{\mathcal{K}} \in \mathbb{R}^{m \times m}$}
	\State $U, S, V^\prime = \mathrm{SVD}(\tilde{\mathcal{K}})$
	\State $\widehat{\mathcal{K}} \gets US^{-\frac{1}{2}}V^\prime$                        \Comment{Normalize $\tilde{\mathcal{K}}$}
	\State $R \gets \mathcal{K}\widehat{\mathcal{K}}$								\Comment{Local topology representation, where $R \in \mathbb{R}^{N \times m}$}
	\State $A_t \gets k\mathrm{NN}(R, k_t)$											\Comment{Contrust $k$NN graph based on $R$}
	\State \Return $G_t = (A_t,\mathbf{X})$
	\end{algorithmic}
\end{algorithm}

\begin{algorithm}[htbp]
	\caption{FT-GCL}
	\label{alg:ft_gcl}
	\begin{algorithmic}[1]
		\State {\bfseries Input:} Graph $G = (V, E, \mathbf{X})$; Upper limit of the number of neighbors $k_{max}$; Pre-generated structural embedding $R$; Number of iterations $T$; Temperature hyper-parameter $\tau$; Hidden and output dimension of the projection head $d$; Hidden dimension and output dimension of the GNN encoder $d^\prime$
		\State $\widehat{A}_f \gets k\mathrm{NN}(\mathbf{X}, k_{max})$               \Comment{Generate $G_f = (\widehat{A}_f, \mathbf{X})$ based on $\mathbf{X}$}
		\State $\widehat{A}_t \gets k\mathrm{NN}(R, k_{max})$                         \Comment{Generate $G_t = (\widehat{A}_t, \mathbf{X})$ based on $R$}
		\State $\mathrm{Use\_FPG} \gets \mathrm{True}$
		\For{$t = 1,2, \cdots, T$}
			\State Random sample $k \leq k_{max}$											
			\If{$\mathrm{Use\_FPG}$ is True}
				\State $A_f = \mathrm{Mask}(\widehat{A}_f, k_{max} -k)$          \Comment{Mask the $k_{max}-k$ least similar neighbors for each node}
				\State $G_f = (A_f,\mathbf{X})$
				\State $G_a \gets G_f$
			\Else
				\State $A_t = \mathrm{Mask}(\widehat{A}_t, k_{max} -k)$
				\State $G_t = (A_t,\mathbf{X})$
				\State $G_a \gets G_t$
			\EndIf
			\State $Z \gets \mathrm{GNN}(G)$, $Z_a \gets \mathrm{GNN}(G_a)$      \Comment{The shared GNN encoder, $Z,Z_a \in \mathbb{R}^{N \times d^\prime}$}
			\State $H \gets g(Z)$, $H_a \gets g(Z_a)$   \Comment{The projection head, $H,H_a \in \mathbb{R}^{N \times d}$}
			\State $\mathcal{L} \gets Eq.(\ref{eq:total_cl})$     \Comment{Channel-level contrastive loss}
			\If{$\mathrm{Use\_FPG}$ is True}                  
				\State $\mathrm{Use\_FPG} \gets \mathrm{False}$
			\Else
				\State $\mathrm{Use\_FPG} \gets \mathrm{True}$
			\EndIf
		\EndFor
	\end{algorithmic}
\end{algorithm}

\section{Reproducibility}
\label{sec:rep}
\subsection{Datasets details}
\subsubsection{Assortative datasets}\label{app:dataset_ass}

The Planetoid~\cite{yang2016revisiting} datasets (i.e., Cora, CiteSeer, and PubMed)  and DBLP~\cite{bojchevski2018deep} are citation networks, where nodes represent documents and edges citation links. WikiCS~\cite{mernyei2020wiki} is a Wikipedia-based dataset, which consists of nodes corresponding to Computer Science articles, and edges based on hyperlinks, as well as 10 classes representing different branches of the field. BlogCatalog~\cite{meng2019co} is a social network with bloggers and their social relationships from the BlogCatalog website, where node features are constructed by the keywords of user profiles. ACM~\cite{wang2019heterogeneous} and CoauthorCS~\cite{sinha2015overview} are coauthor networks, where edges' endpoints are two papers with the same authors.

\subsubsection{Disassortative datasets}\label{app:dataset_disass}

Texas, Wisconsin and Cornell are graphs where nodes are web pages of the corresponding universities and edges are links between web pages. Node features are the bag-of-words representation of web pages. These graphs are originally collected by the CMU WebKB project. Actor is a co-occurence network where each node correspond to an actor, and the edge between two nodes denotes co-occurrence on the same Wikipedia page. Node features correspond to some keywords in the Wikipedia pages. We used the preprocessed version of these four datasets in~\cite{pei2019geom}.

\subsection{Implementation Details}\label{app:imp_det}

In the graph view generation stage, for topology proximity graph $G_t$ the local subgraph of each node is induced by $\gamma$ random walk sequences of length $l$. We use alias sampling such that the sampling complexity is $\mathcal{O}(\gamma l N)$. We set $\gamma=30$, $l = 10$ for all benchmark graphs. We use the WL kernel with the Nyström method to encode local topology for each node, and the number of basis vectors $m$ is set to $200$ for all graphs. For the shared GNN encoder, we use 2-layers Graph Attention Networks~\cite{velickovic2018graph} with a single head by default. FT-GCL does not impose any constraint on the GNN architecture, so other GNNs such as GraphSAGE can be also used. The activation function $\sigma$ of the GNN encoder is selected from $\{\mathrm{ReLU}, \mathrm{eLU}, \mathrm{PReLU}\}$. We use Adam~\cite{kingma2014adam} for optimizing $\mathcal{L}_{cont}$ with learning rate $lr$ from $\{1e^{-2},1e^{-3},5e^{-3}, 5e^{-4}\}$. The hidden layer and the output layer of the GNN encoder have the same number of dimensions $d^\prime$, which is selected from $\{64,128,256,512\}$. The hidden layer and the output layer of the MLP have the same number of dimensions $d$, which is also selected from $\{64,128,256,512\}$. The value of weight decay is selected from $\lambda \in \{1e^{-4}, 5e^{-3},5e^{-4}.5e^{-5}\}$
The experiments are implemented using Pytorch Geometric 1.6.1~\cite{fey2019fast} with CUDA version 10.2, networkx 2.5, Python 3.8.5, scikit-learn~\cite{pedregosa2011scikit} 0.23.2. The experiments are conducted on Linux servers installed with a NVIDIA Quadro RTX8000 GPU and 10 Intel(R) Xeon(R) Silver 4210R CPUs.

\subsection{Hyperparameter Specifications}\label{app:hyper}

\begin{table}[h!]
	\begin{center}
		\caption{Hyperparameter for FT-GCL on all benchmarks.}
		\label{tab:hyper}

			\begin{tabular}{ccccccccc} 
				\toprule			
				&Benchmarks & $k_{max}$ & $d$ & $d^\prime$ & $lr$      & $\tau$ & $\lambda$ &$\sigma$\\
				\midrule
				&Cora  		&     8     & 512 & 256        & $5e^{-4}$ & 0.2    & $5e^{-5}$ & PReLU\\
				&CiteSeer   &     5     & 512 & 64         & $5e^{-4}$ & 0.2    & $5e^{-3}$ & eLU\\
				&PubMed     &     10    & 512 & 128        & $5e^{-3}$ & 0.9    & $5e^{-4}$ & eLU\\
				&DBLP		&     10    & 512 & 64         & $5e^{-4}$ & 0.4    & $5e^{-5}$ & PReLU\\
				&WikiCS		&     10    & 256 & 128        & $5e^{-4}$ & 0.2    & $5e^{-4}$ & eLU\\
				&ACM		&     10    & 512 & 64         & $1e^{-3}$ & 0.7    & $5e^{-4}$ & ReLU\\
				&CoauthorCS &     10    & 512 & 512        & $5e^{-4}$ & 0.1    & $1e^{-4}$ & eLU\\
				&Texas		&     10    & 64  & 128        & $1e^{-3}$ & 0.2    & $5e^{-4}$ & ReLU\\
				&Wisconsin	&     10    & 64  & 128        & $1e^{-3}$ & 0.2    & $5e^{-4}$ & ReLU\\
				&Cornell	&     10    & 128 & 128        & $1e^{-3}$ & 0.2    & $5e^{-4}$ & ReLU\\
				&Actor      &     10    & 128 & 64         & $1e^{-2}$ & 0.2    & $5e^{-5}$ & ReLU\\
				\bottomrule
		\end{tabular}
	\end{center}
\end{table}

\section{Analysis of \Cref{sec:disassortative}}
\label{app:more_ana}
The results of node classification task on disassortative graphs in \Cref{tab:hetero} show that FT-GCL and its two variants F-GCL and T-GCL outperform other models with a significant margin. In this section, we give an analysis based on the edge homophily ratio (EHR) defined in~\citet{zhu2020beyond}:
\begin{equation}
	h=\frac{\left|\left\{(v_i, v_j):(v_i, v_j) \in E \wedge y_{i}=y_{j}\right\}\right|}{|E|},
\end{equation}
which means the proportion of the intra-class edges to the total number of edges. It can be used as a measure of the graph's homophily level. For all disassortative benchmarks, we report the EHR values on the original graph $G$, its FPG $G_f$ and TPG $G_t$ in \Cref{tab:homo_ratio}. We find that all the four datasets have weak homophily. However, both the FPG and TPG of Texas and Cornell have strong homophily, i.e., $h_f, h_t > 0.5$. This explains why F-GCL, T-GCL and FT-GCL outperform other methods. Since Wisconsin's FPG has strong homophily, F-GCL and FT-GCL achieve the best results. It is worth noting that although the two proximity graphs of Actor have weak homophily, exploiting multiple types of proximity can also help our model learn relatively better embeddings. The experiments in \Cref{sec:disassortative} suggest that FPG and TPG are complementary to the original graph, which is particularly evident in the disassortative graphs with low homophily.

\begin{table}[h!]
	\begin{center}
		\caption{The edge homophily ratio of $G$, $G_f$ and $G_t$ ($k_f = k_s = 6$) on four disassortative graphs.}
		\label{tab:homo_ratio}
		\begin{tabular}{ccccccc}
			\toprule
			&\textbf{Texas} & \textbf{Wisconsin} & \textbf{Cornell} & \textbf{Actor}\\
			\midrule
			Homo. ratio $h$			&0.108 & 0.196 & 0.305  & 0.219\\
			FPG Homophily ratio $h_f$	&0.657 & 0.699 & 0.657	& 0.250\\
			TPG Homophily ratio $h_t$	&0.530 & 0.337 & 0.530	& 0.233\\
			\bottomrule
		\end{tabular}
	\end{center}
\end{table}

\section{More experiments}
\label{app:more_exp}

\subsection{Node classification}
In this section, we show the results of node classification for the DBLP, WikiCS, BlogCatalog, ACM and CoathorCS datasets in \Cref{tab:more_exp}. Among these four additional experiments, the previous state-of-the-art GRACE and MVGRL achieve the best results on DBLP and BlogCatalog, respectively. Our proposed FT-GCL performs the best on WikiCS and ACM, and achieves the second-best results on CoauthorCS among all unsupervised methods. In addition, F-GCL achieves the second-best results on DBLP, WikiCS and ACM, and is the best unsupervised method on CoauthorCS. 


\begin{table}[h!]
	\centering
	\caption{Classification accuracies(\%) and macro-F1 scores(\%) with standard deviation.\crule[green]{0.3cm}{0.3cm} is the champion and \crule[red!30!white!100]{0.3cm}{0.3cm} is the runner-up.}
	\label{tab:more_exp}
	\scalebox{0.5}{
		\begin{tabular}{c|c|c|cc|cc|cc|ccccccc} 
			\toprule			
			&\multirow{2}{*}{\textbf{Input}}&\multirow{2}{*}{\textbf{Algo.}}& \multicolumn{2}{c|}{\textbf{DBLP}} & \multicolumn{2}{c|}{\textbf{WikiCS}}& \multicolumn{2}{c|}{\textbf{ACM}} & \multicolumn{2}{c}{\textbf{CoauthorCS}}\\
			&			   &				&Acc&Macro-F1 &Acc&Macro-F1					&Acc&Macro-F1 &Acc&Macro-F1 \\
			\midrule
			\multirow{15}{*}{\textbf{Unsupervised}}		&\multirow{1}{*}{$X$} 					&Row features 		& $71.22(\pm0.71)$& $64.62(\pm0.76)$&$72.67(\pm0.62)$&$69.24(\pm0.82)$&$86.40(\pm0.42)$&$86.40(\pm0.41)$&$90.18(\pm0.07)$&$87.21(\pm0.14)$\\  \cmidrule(l){2-11}
			&\multirow{4}{*}{$A$} 					&DeepWalk 			& $78.92(\pm0.20)$ &$72.71(\pm0.80)$&$68.33(\pm1.27)$&$64.64(\pm1.39)$&$60.09(\pm0.16)$&$55.32(\pm1.71)$&$85.00(\pm0.38)$&$83.58(\pm0.29)$\\  
			&    									&node2vec 			&$79.45(\pm0.24)$&$73.11(\pm0.36)$&$69.06(\pm0.75)$&$65.19(\pm0.80)$&$61.15(\pm0.77)$&$57.28(\pm0.09)$&$84.97(\pm0.36)$&$83.29(\pm0.62)$ \\ 
			&    									&LINE     			& $64.03(\pm0.33)$&$56.45(\pm0.70)$&$67.36(\pm0.89)$&$63.84(\pm0.96)$&$41.33(\pm0.57)$&$39.90(\pm0.25)$&$77.45(\pm0.40)$&$69.95(\pm1.23)$\\ 
			&										&VERSE     			&$79.22(\pm0.33)$&$73.07(\pm0.58)$&$72.67(\pm0.62)$&$69.24(\pm0.82)$&$69.18(\pm0.83)$&$69.16(\pm0.83)$&$86.56(\pm0.30)$&$82.75(\pm\textbf{\textbf{}}0.68)$\\ \cmidrule(l){2-11}
			&\multirow{10}{*}{$A$,$X$} 				&GAE      			& $82.36(\pm0.18)$ &$78.91(\pm0.39)$ &$68.72(\pm0.52)$ &$60.84(\pm0.90)$&$87.62(\pm0.66)$& $87.51(\pm0.65)$ &$92.46(\pm0.20)$&$90.09(\pm0.44)$\\
			&										&VGAE				& $82.68(\pm0.22)$ &$79.45(\pm0.34)$ &$75.64(\pm0.63)$&$71.83(\pm0.84)$&$89.78(\pm0.48)$&$89.73(\pm0.47)$&$92.10(\pm0.22)$&$89.85(\pm0.46)$\\
			&										&ARVGA				& $75.77(\pm0.43)$&$66.83(\pm0.85)$&$67.35(\pm0.49)$&$60.75(\pm0.83)$&$74.07(\pm1.31)$& $74.37(\pm1.48)$&$88.54(\pm0.23)$&$83.76(\pm0.38)$\\
			&										&MVGRL				& $83.68(\pm0.17)$&$79.54(\pm0.24)$&$74.13(\pm0.64)$&$68.89(\pm1.42)$&$90.80(\pm0.31)$&$90.81(\pm0.32)$&$92.08(\pm0.50)$&$90.93(\pm0.42)$\\
			&										&DGI      			& $83.11(\pm0.19)$&$78.64(\pm0.44)$&$75.70(\pm0.45)$&$72.12(\pm0.69)$&$90.01(\pm0.30)$&$90.00(\pm0.33)$&$92.30(\pm0.16)$&$90.45(\pm0.32)$\\
			&										&GRACE    			&\cellcolor{green} $\mathbf{84.45}(\pm0.19)$&\cellcolor{green}$\mathbf{80.31}(\pm0.26)$ &$78.26(\pm0.15)$&$75.96(\pm0.71)$&$90.28(\pm0.46)$&$90.27(\pm0.48)$&$92.42(\pm0.20)$&$90.44(\pm0.25)$\\
			&										&GMI-Adaptive		& $83.51(\pm0.32)$&$78.66(\pm0.40)$ &$77.57(\pm0.44)$&$74.65(\pm0.63)$&$87.15(\pm0.53)$&$87.17(\pm0.53)$ &$91.92(\pm0.13)$&$89.06(\pm0.31)$\\		\cmidrule(l){3-11}
			&										&\textbf{F-GCL}		&\cellcolor{red!30!white!100} $83.85(\pm0.15)$&\cellcolor{red!30!white!100}$79.74(\pm0.21)$ &\cellcolor{red!30!white!100}$78.69(\pm0.66)$&\cellcolor{red!30!white!100}$75.38(\pm0.70)$&\cellcolor{red!30!white!100}$90.99(\pm0.15)$ &\cellcolor{red!30!white!100}$91.01(\pm0.15)$&\cellcolor{green}$\mathbf{93.06}(\pm0.15)$&\cellcolor{red!30!white!100}$91.04(\pm0.20)$\\
			&										&\textbf{T-GCL}     & $82.79(\pm0.35)$&$78.43(\pm0.55)$&$76.74(\pm0.23)$&$72.81(\pm0.15)$&$87.39(\pm0.33)$&$87.44(\pm0.33)$&$92.77(\pm0.04)$&$90.74(\pm0.27)$\\
			&										&\textbf{FT-GCL}	&  $83.58(\pm0.64)$&$79.52(\pm0.39)$&\cellcolor{green}$\mathbf{78.91}(\pm0.61)$&\cellcolor{green}$\mathbf{76.53}(\pm0.38)$&\cellcolor{green}$\mathbf{91.28}(\pm0.18)$&\cellcolor{green}$\mathbf{91.28}(\pm0.19)$&$92.94(\pm0.03)$&$90.81(\pm0.27)$\\
			\midrule
			\multirow{4}{*}{\textbf{Semi-supervised}}	&\multirow{4}{*}{$A$,$X$,$Y_{train}$}  	&GCN 				&$82.73(\pm0.58)$&$77.34(\pm0.29)$&$76.40(\pm1.00)$&$72.35(\pm0.21)$ &$87.66(\pm1.18)$&$87.76(\pm1.51)$ &\cellcolor{red!30!white!100}$93.05(\pm0.26)$&\cellcolor{green}$\mathbf{91.07}(\pm0.22)$\\ 
			&									  	&GAT 				&$83.39(\pm0.16)$&$79.11(\pm0.10)$ &$77.60(\pm0.60)$&$74.37(\pm0.88)$&$86.52(\pm0.97)$&$86.75(\pm1.08)$ &$92.42(\pm0.29)$&$90.38(\pm0.10)$\\
			&									  	&SGC 				&$82.10(\pm0.12)$&$76.99(\pm0.56)$ &$74.92(\pm0.22)$&$71.66(\pm0.23)$&$88.65(\pm0.37)$&$88.77(\pm0.26)$&$93.10(\pm0.78)$&$90.88(\pm0.30)$\\
			&									  	&JK-Net 			&$81.82(\pm0.80)$&$77.54(\pm0.35)$ &$68.93(\pm0.89)$&$66.75(\pm0.62)$&$88.20(\pm0.80)$&$88.23(\pm0.16)$&$89.60(\pm0.22)$&$85.96(\pm0.10)$\\
			
			\bottomrule
		\end{tabular}
	}
\end{table}

We also conduct experiments on public splits of Planetoid datasets provided by~\cite{yang2016revisiting}, and report the results in \Cref{tab:public_split}.

\begin{table}[h!]
	\centering
	\caption{Averaged classification accuracies(\%) with standard deviation on Planetoid datasets (public splits).\crule[green]{0.3cm}{0.3cm} is the champion and \crule[red!30!white!100]{0.3cm}{0.3cm} is the runner-up.}
	\label{tab:public_split}
	\scalebox{0.85}{
	\begin{tabular}{c|c|c|ccccccccc} 
		\toprule			
		&\textbf{Input}							&\textbf{Algo.}	&\textbf{Cora} &\textbf{CiteSeer}&\textbf{PubMed}\\
		\midrule
		&\multirow{10}{*}{$A$,$X$} 				&GAE	                                &$71.5(\pm0.4)$&$65.8(\pm0.4)$&$72.1(\pm0.5)$                                \\
		\multirow{10}{*}{\textbf{Unsupervised}}& &VGAE                                  &$75.2(\pm0.2)$&$69.0(\pm0.2)$&$75.3(\pm0.5)$	                            \\
		&										&ARVGA		                            &$74.4(\pm0.1)$&$64.2(\pm0.6)$&$74.7(\pm0.7)$                                    \\
		&										&MVGRL     			                    &$82.9(\pm0.7)$&\cellcolor{red!30!white!100}$72.6(\pm0.7)$&$79.4(\pm0.3)$                                     \\
		&										&DGI      	                            &$82.3(\pm0.6)$&$71.8(\pm0.7)$&$76.8(\pm0.6)$                                           \\
		&										&GRACE    		                        &$80.0(\pm0.4)$&$71.7(\pm0.6)$&$79.5(\pm1.1)$                                         \\
		&										&GMI-Adaptive	                        &\cellcolor{red!30!white!100}$83.0(\pm0.3)$&$72.4(\pm0.1)$&\cellcolor{green}$\mathbf{79.9}(\pm0.2)$                                          \\ \cmidrule(l){3-6}
		&										&\textbf{F-GCL}	&$82.9(\pm0.5)$&$72.2(\pm0.4)$&\cellcolor{green}$\mathbf{79.9}(\pm0.3)$	                                                                 \\
		&										&\textbf{T-GCL} &$80.8(\pm0.1)$&\cellcolor{green}$\mathbf{73.3}(\pm0.2)$&$77.7(\pm0.7)$                                                                  \\
		&										&\textbf{FT-GCL}	                   &\cellcolor{green}$\mathbf{83.1}(\pm0.3)$&\cellcolor{red!30!white!100}$72.6(\pm0.3)$&\cellcolor{red!30!white!100}$79.8(\pm0.6)$	\\
		\midrule
		\multirow{3}{*}{\textbf{Semi-supervised}}	&\multirow{4}{*}{$A$,$X$,$Y_{train}$}  	&GCN&$81.5(\pm0.4)$&$70.3(\pm0.7)$ &$79.0(\pm0.5)$\\ 
		&									  	                                            &GAT&\cellcolor{red!30!white!100}$83.0(\pm0.7)$&$72.5(\pm0.7)$&$79.0(\pm0.3)$  \\
		&									  	                                            &SGC&$81.0(\pm0.0)$&$71.9(\pm0.1)$&$78.9(\pm0.0)$\\
		\bottomrule
	\end{tabular}
	}
\end{table}

\subsection{Link Prediction}
In this section, we evaluate the performance on the link prediction task. We conduct experiments on Cora and CiteSeer, removing 5\% existing edges and the same-size non-existing edges for validation, 10\% existing edges and the same-size non-existing edges for testing. In this task, we need to compute the similarity between node embeddings of GNN encoder directly, so we remove the projection head of our model. We adopt an inner product decoder to obtain the predicted adjacency matrix:
\begin{equation}
	A_{pred} =  \sigma(ZZ^\top)
\end{equation}
where $\sigma$ is the sigmoid function. We use 10 random train/validation/test splits and run 10 times on each split. We report the AUC and AP in percentage with top-2 highlighted in \Cref{tab:lp}. The F-GCL and FT-GCL achieve the highest prediction scores among all baseline models on CiteSeer and Cora, indicating that our model has better reasoning ability for network connectivity and can better retain similarity information between the nodes.
\begin{table}[h!]
	\begin{center}
		\caption{Experimental results of link prediction.\crule[green]{0.3cm}{0.3cm} is the champion and \crule[red!30!white!100]{0.3cm}{0.3cm} is the runner-up.}
		\label{tab:lp}

		\begin{tabular}{cc|cc|ccccc} 
			\toprule
			
			&\multirow{2}{*}{\textbf{Methods}} & \multicolumn{2}{c|}{\textbf{Cora}} & \multicolumn{2}{c}{\textbf{CiteSeer}}\\
			&								   & AUC 					& AP       &AUC & AP\\   \midrule
			&GAE							   &$91.37(\pm0.56)$		  &$91.99(\pm0.69)$ &$89.53(\pm0.66)$&$92.02(\pm0.83)$									\\
			&VGAE							   &$92.08(\pm0.86)$     	  &\cellcolor{red!30!white!100}$92.60(\pm0.02)$&$90.68(\pm1.04)$&$92.17(\pm0.18)$									\\
			&ARVGA							   &$92.40(\pm1.08)$		  &$92.39(\pm0.83)$&$92.51(\pm0.81)$&$93.02(\pm0.82)$											\\
			&DGI							   &$91.19(\pm1.11)$		  &$91.46(\pm1.01)$	&$90.36(\pm1.19)$&$90.16(\pm1.51)$											\\
			&GCN							   &$91.52(\pm0.85)$      	  &$91.56(\pm1.00)$&$90.00(\pm1.56)$&$90.90(\pm1.18)$	 							    \\
			\midrule
			&\textbf{F-GCL}					   &\cellcolor{red!30!white!100}$92.97(\pm0.39)$      &$92.46(\pm0.64)$&\cellcolor{green}$\mathbf{95.24}(\pm0.62)$&\cellcolor{green}$\mathbf{95.06}(\pm0.71)$&												\\
			&\textbf{T-GCL}					   &$89.95(\pm1.03)$		&$90.16(\pm1.11)$&$93.64(\pm0.70)$&$93.93(\pm0.72)$&											\\
			&\textbf{FT-GCL}				   &\cellcolor{green}$\mathbf{92.98}(\pm0.81)$		&\cellcolor{green}$\mathbf{92.62}(\pm0.80)$&\cellcolor{red!30!white!100}$95.09(\pm0.67)$&\cellcolor{red!30!white!100}$94.95(\pm0.59)$&							\\
			\bottomrule
		\end{tabular}

	\end{center}
\end{table}

%

\subsection{Sensitivity analysis}
\begin{figure}[htbp]
	\centering
	\includegraphics[scale=0.5]{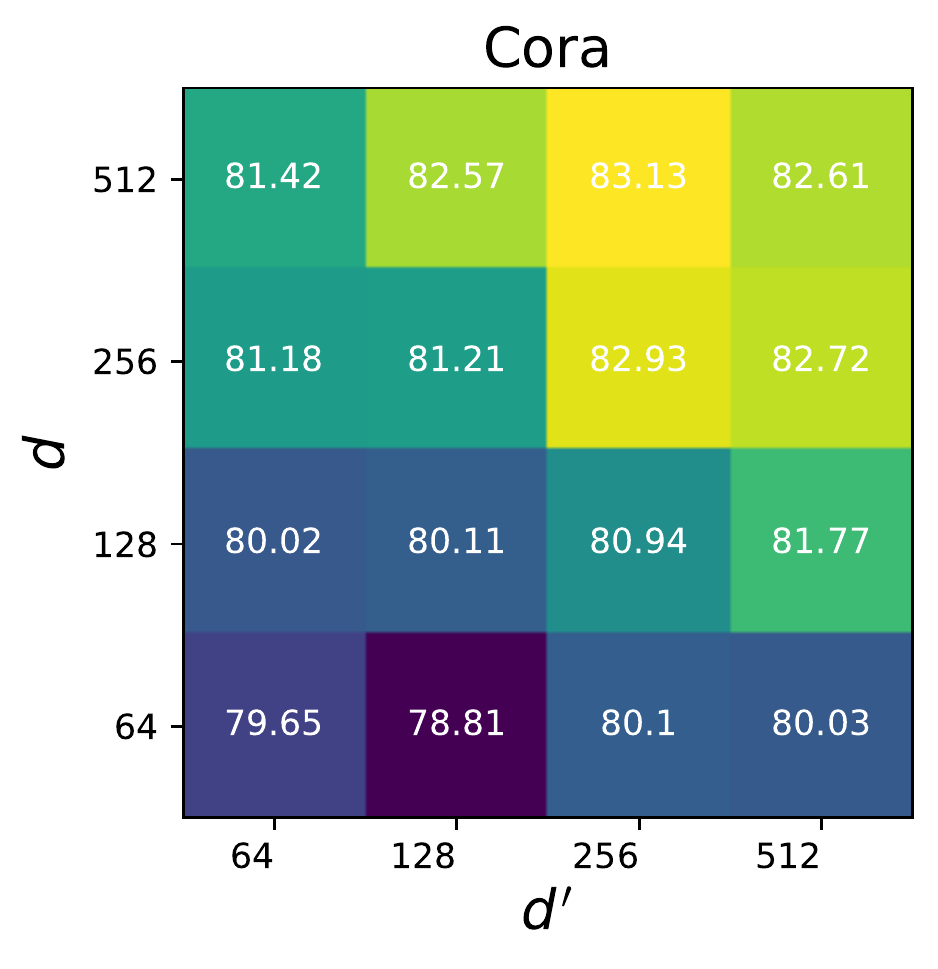}
	\hspace{1in}
	\includegraphics[scale=0.5]{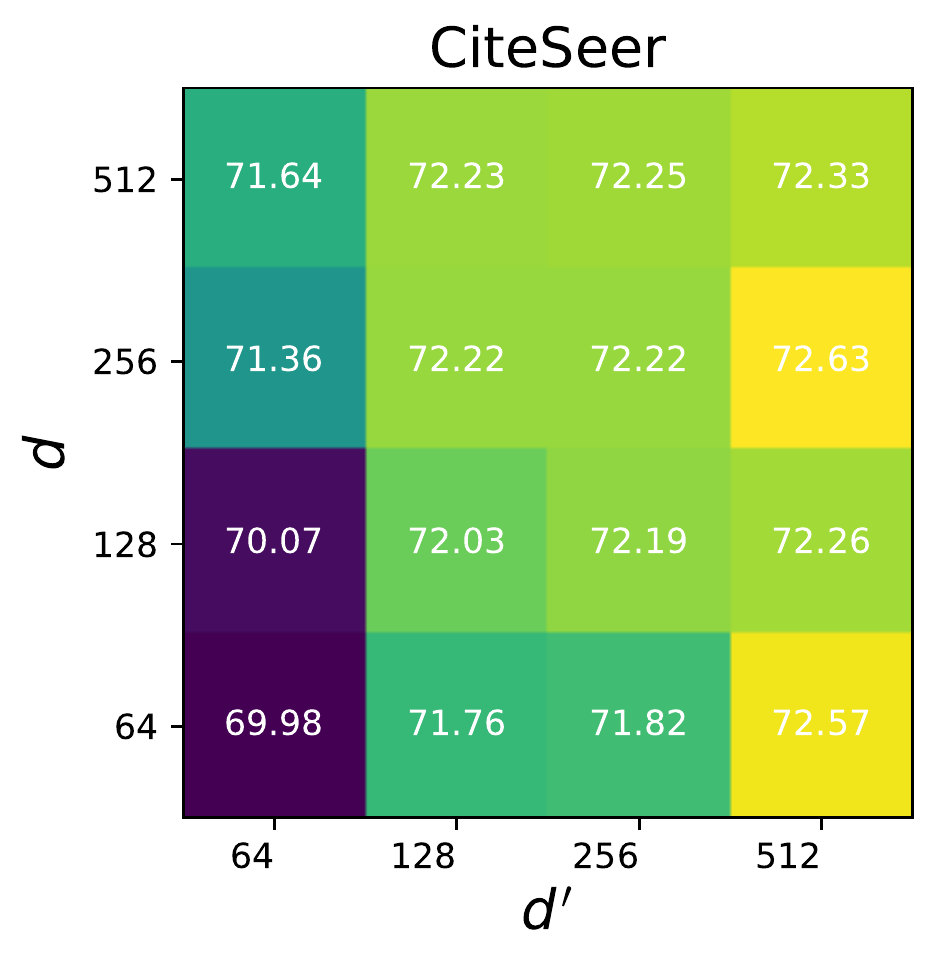}
	\caption{Parameter sensitivity in node classification.}
	\label{fig:sen}
\end{figure}

\Cref{fig:sen} shows the impact of the number of dimensions for the hidden layers of the GNN encoder and the MLP on the node classification performance of FT-GCL on Cora and CiteSeer. With the increase of dimensionalities, the performance increases first and tends to stability when the dimensionalities are 256 and 512 respectively. We can also see that when both dimensionalities take the maximum, performances are sub-optimal. Besides, the two heatmaps show a similar trend.  

\end{document}